\pdfoutput=1
\documentclass[11pt]{article}

\usepackage{ACL2023}

\usepackage{times}
\usepackage{latexsym}
\PassOptionsToPackage{usenames, dvipsnames}{color}

\usepackage[T1]{fontenc}
\usepackage[utf8]{inputenc}
\usepackage{wrapfig,lipsum,booktabs}
\usepackage{subcaption}
\usepackage{microtype}
\usepackage{array,multirow}
\usepackage{xspace}
\usepackage[inline]{enumitem}

\usepackage{inconsolata}
\usepackage{graphicx}
\usepackage{amsmath}
\usepackage{amsthm}
\usepackage{amssymb} 
\usepackage{bm} 
\usepackage{array}
\newcolumntype{H}{>{\setbox0=\hbox\bgroup}c<{\egroup}@{}}
\newtheorem*{unnumberedtheorem}{Theorem}
\newcommand\stdfine{\texttt{Direct}\xspace}
\newcommand\removetoken{\texttt{RemoveToken}\xspace}
\newcommand\dfl{\texttt{DFL}\xspace}
\newcommand\dflsuccess{\texttt{DFL-nodemog}\xspace}
\newcommand\poe{\texttt{POE}\xspace}
\newcommand\inlp{\texttt{INLP}\xspace}
\newcommand\subsample{\texttt{Subsample}\xspace}
\newcommand\groupdro{\texttt{GroupDRO}\xspace}
\newcommand\riesz{\texttt{Riesz}\xspace}
\newcommand\propen{\texttt{Propensity}\xspace}

\newcommand{\cer}[1]{\texttt{FEAG}({#1})}
\newcommand{\cero}{\texttt{FEAG}\xspace}
\newcommand\rmdo{\textrm{do}}
\newcommand\ate{\textrm{ATE}}

\newcommand\treated{\texttt{Treated}\xspace}
\newcommand\untreated{\texttt{Untreated}\xspace}
\newcommand\dir{\textrm{Direct}}
\newcommand\dr{\textrm{DR}}
\newcommand\pr{\textrm{Pr}}
\newcommand\rr{\textrm{R}}

\usepackage[color=lightgray]{todonotes}
\title{Controlling Learned Effects  to Reduce Spurious Correlations in Text Classifiers}

\author{Parikshit Bansal \\
  Microsoft Research, India \\
  \texttt{parikshitb52@gmail.com} \\\And
  Amit Sharma \\
  Microsoft Research, India\\
  \texttt{amshar@microsoft.com} \\}

\begin{document}
% \nolinenumbers
\maketitle

\begin{abstract}
% writing for first few days, then get into experiments 0.5. First on writing. 
% writing focus. 

% Previous works just removes. 
% causal effect is non-zero screw up id. 
%To address the problem of spurious correlations learnt by NLP classifiers  between training features and target labels . 
To address the problem of NLP classifiers learning  spurious correlations between training features and target labels, a common approach is to make the model's predictions invariant to these features. However, this can be counter-productive when the features have a \emph{non-zero} causal effect on the target label and thus are important for prediction. Therefore, % the correct objective is to regularize the model's learned causal effect based on the causal effect of the features on target label. % with a strength that depends on the desired tradeoff between in-distribution and out-of-distribution accuracy/average accuracy and minority groups' accuracy.  %that  Prior work solves this by removing these correlations using invariance based learning constraints, making model predictions indifferent to these features. But often these spurious features are important for prediction, having a non-zero causal effect on the target label. Learning models invariant to these features can hurt model's accuracy on (all distributions?)
% therefore we provide finegrained control, set the learn effect to some value maximising ID accuracy while debiasing as much possible depending on the tradeoff between IID and OOD accuracy.
using methods from the causal inference literature, 
we propose an algorithm to regularize the learnt effect of the features on the model's prediction to the estimated effect of feature on label.  % or to a value which minimises the spurious correlations while improving both average and minority groups (i.e. samples breaking spurious correlations) accuracies. 
% do so by using method from causal effect literature, that help us estimate effect of token on dataset label. 
%We use methods from causal effect estimation literature to estimate the effect of a feature on the target label. 
% regularise/augment our main classifier. can be seen as automated data augmentation method. 
%This value can then be used to regularise the importance of the features. 
This results in an automated augmentation method that leverages the estimated effect of a feature to appropriately change the labels for new augmented inputs. %text. 
% 1-2 lines about results
On toxicity and IMDB review datasets, the proposed algorithm minimises spurious correlations and improves the minority group (i.e., samples breaking spurious correlations) accuracy, while also improving the total accuracy compared to standard training. % tradeoff between average group and minority group accuracies.
% in addition to training debiased classifier, its useful in its own right. Use to detect annotator bias, when true causal effect of tokens is known. 
% We also show the usefulness of our method for detecting annotator bias in training data.
\footnote{\footnotesize{Code: \url{https://github.com/pbansal5/feature-effect-augmentation}}}%, whenever the true causal effect of an input feature is known. 
\end{abstract}

\section{Introduction}
\label{sec:intro}

% \todo{Example for it. Our pitch. Introduce reisz. Citations, good claim on why removing is bad. good pitch}

%% Introduction
While classifiers trained on pre-trained NLP models achieve state-of-the-art accuracy on various tasks, they have been shown to learn spurious correlations between input features and the label~\cite{du2022shortcut}. Such learned correlations impact accuracy on out-of-distribution samples and in the case of \textit{sensitive} spurious features, lead to unfair predictions~\cite{sun2019mitigating,ribeiro2020beyond}.  
Learned spurious correlations can be over features that are either irrelevant (e.g., tense, gender for profession classification) or relevant (e.g., emoticons for sentiment classification, negation words for  contradiction). In both cases, the classifier overweighs their importance compared to other features. % and necessary   These apart from making models unsafe for deployment also hurt their in-domain and minority groups i.e. underrepresented groups accuracies. These minority groups break the spurious correlation prevalent in the dataset leading to model's degraded performance on these samples. From user demographic shifts in hate speech classifiers to catalogue changes in text-based recommendation systems, these shifts are common in text data.
%\amit{need to specify that we are talking about NLP classifiers based on pre-trained models. Also this is too early to define minority group. we should start with examples and cites of biases corr. in NLP models.}

% Prior solution of removing them
For removing  spurious correlations, a common principle underlying past work is to make a model's prediction \textit{invariant} to the features that exhibit the correlation. This can be done by  data augmentation \cite{kaushik2019learning}, latent space removal \cite{ravfogel2020null}, subsampling \cite{sagawa2019distributionally,sagawa2020investigation}, or sample reweighing \cite{mahabadi2019end,orgad2022debiasing}. 
In many cases, however, the correlated features may be important for the task and their complete removal can cause a degradation in task performance. For instance, for spurious correlation over negation tokens (e.g., ``not'') or lexical overlap in MNLI natural language inference tasks, \citet{williams2017broad,joshi2022all}   show that correlated features  are necessary for prediction and their removal can hurt accuracy. %\amit{need to give some context, a few phrases, to describe what is negation bias, what is lexical bias, what is MNLI}

%% Example of why removing is bad

\begin{figure}[t]
\centering
\includegraphics[width=0.4\textwidth]{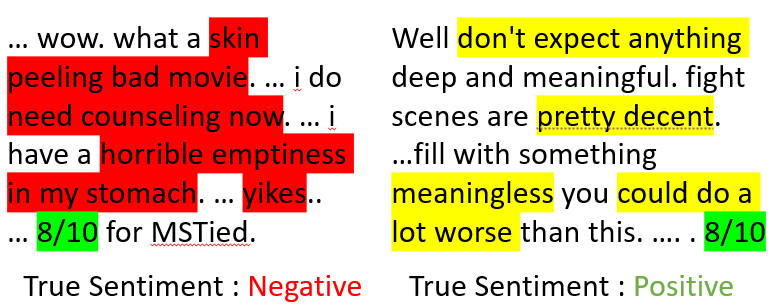}
\caption{Example from IMDB reviews dataset showing the spurious token ``8/10'' and its importance for prediction on some inputs. Parts highlighted in \textcolor{orange!40!yellow}{yellow are \textbf{ambiguous}} in sentiment, in \textcolor{green}{green are (supposedly) \textbf{positive}} in sentiment and \textcolor{red}{red are \textbf{negative}}.}
\label{fig:eg}
\vspace*{-15pt}
\end{figure}

As another example, consider the IMDB review dataset~\cite{maas-EtAl:2011:ACL-HLT2011} where the task is classify the sentiment of a given review as positive or negative. Reviewers often include a numeric rating in their text reviews, e.g., \texttt{``9/10''} or \texttt{``1/10''}. The numeric rating is highly correlated with the sentiment label, often regarded as a  spurious correlation~\cite{pezeshkpour2021combining} that a model should not rely on. %At the same time, removing the numeric rating from the input can be an over-reaction since the rating may contain valuable information about the reviewer's sentiment. %Their occurrence is highly correlated with the target task of sentiment classification with  \texttt{9/10} suggesting that the reviewer has a positive sentiment while \texttt{1/10} suggesting a negative sentiment. 
%While these tokens are important for the task (the percieved sentiment is affected by the rating given by the reviewer), they are not sufficient for the task (a reviewer might include \texttt{9/10} to praise certain aspects of the movie, while having a generally negative sentiment). 
In the first review of Fig.~\ref{fig:eg}, for instance, the positive rating can mislead a classifier since the review is overall negative. However, in the second example,  the text is ambiguous and the rating \texttt{``8/10''}  can provide a helpful signal about the reviewer's sentiment (and removing it may decrease classifier's accuracy). Thus, there exist inputs where the rating is a helpful feature for prediction and other inputs where it can be counter-productive. %This shows that heavily relying on these strings can be harmful as well. 
This shows the trade-off between accuracy on \emph{majority} groups, (i.e.,  samples where these correlations hold and constitute a majority of samples) and \emph{minority} groups (i.e., comparatively fewer samples where these correlations break).

%Rather than completely removing the effect of a feature on model's prediction, we propose that 
In this paper, we propose a general method to resolve the above trade-off: rather than always removing the effect of a feature on the model's prediction, we argue that the learned effect should be equal to the \textit{true effect of the feature} on the output label. We define feature effect using the notion of conditional effect from the causal inference literature~\cite{pearl2009causality}: the  change in the ground-truth label upon changing the feature, keeping all other input features constant. %Thus, the learned effect can be zero for  irrelevant features but not need not be for other relevant features. 
To enforce the true feature effect, we make \textbf{two contributions}: 
\begin{enumerate}%[leftmargin=*,topsep=0.5pt,noitemsep]{}
\item Novel estimator of the effect of text features on the label that is accurate even at high levels of spurious correlation compared to past work.
\item Automated augmentation method that predicts the labels of new samples using the estimated feature effect and adds them to train data to achieve the desired learned effect in a classifier.
\end{enumerate}
%\todo{\amit{better way to motivate: what if we change 1/10 to 9/10 for a bad review--classifier should not flip its prediction label only based on 9/10. }}

%The task of estimating causal effect is hard. we provide novel estimator.
%augmentation for improving classifier.
When combined with the standard accuracy loss over training data, the proposed method, \textit{Feature Effect Augmentation (FEAG)}, obtains the highest overall accuracy compared to baselines while reducing the learnt spurious correlation. For our evaluation, we consider the practical goal of increasing the accuracy on the minority groups while not substantially reducing the accuracy over the majority group.   On comment toxicity and IMDB review datasets, we find that existing methods tend to increase minority group accuracy but reduce overall accuracy, whereas FEAG obtains a good tradeoff. In some cases, it can obtain both  higher overall accuracy and higher average group accuracy. %That said, maximizing overall accuracy may not always be the desired goal. 
Moreover, by making it easy to change the  target feature effect to be enforced,   FEAG provides an interpretable control mechanism to obtain any  desired tradeoff between minority and majority group accuracy (setting the feature effect to zero, e.g.,  prioritizes minority group accuracy). %. one may want to maximize the average accuracy across groups or another tradeoff between overall and group  accuracy. The proposed algorithm FEAG allows a  fine-grained, interpretable control over the model's performance: by changing the target learned effect, one can obtain a classifier with the desired tradeoff between overall and group accuracy. Setting the target learned effect to zero, for example, corresponds to removing a feature's effect as in past work and would lead to the highest minority group accuracy. 

More generally, our work provides a viable direction for automated data augmentation. While existing work requires manual labeling of counterfactual examples for removing spurious correlation~\cite{kaushik2019learning,wu2021polyjuice}, our method can label new examples using estimated feature effects. We also show how estimated feature effects can be useful for other tasks, such as detecting annotator bias in a train set. 

\section{Related Work}
\label{sec:related}
%\pari{all methods aim to go to 0}
%\pari{interpretable thing. given a goal ATE we can get you that.}
Our work combines the debiasing NLP literature with causal effect estimation over text. %cusing on removal of a biasing attribute/spurious feature.

%\amit{divide related work by the different kinds of debiasing methods. Our goal is to show that most NLP focuses on removal of spurious feature}
%\amit{Have a named para for null space removal, weighting methods, cf augmentation. In the third para, you can talk about how our method augments based on effect, not to zero effect}

\subsection{Estimating causal effect from text}
%\paragraph{Causal Effect Estimation} Propensity estimation in high dimensional spaces like text data is notoriously hard.
Prior work on estimating causal effect on text is based on propensity scores, such as DragonNet \cite{shi2019adapting} and follow-up work~\cite{veitch2020adapting,gui2022causal}. However, propensity-based estimators are known to suffer from high variance, especially in text scenarios where overlap may be low~\cite{gui2022causal}. We utilize a Riesz-based causal estimator \cite{chernozhukov2022riesznet} that has recently been shown to offer a better bias-variance tradeoff. In particular, it does not need to estimate the full propensity  but rather estimates the weight for each sample directly, thus avoiding the variance issues of prior methods. % a reliable method for estimating causal effect. % tries to solve this by modeling inverse of propensities directly instead of modeling propensities and plugging them into the inverse weighing analytical formula. 

\subsection{Removing spurious correlations }
%There is a rich literature on removal of biasing attribute/spurious feature. 

\noindent \textbf{Latent Space Removal.} These methods aim to remove the spurious feature from model's learnt representation. INLP \cite{ravfogel2020null} removes  spurious features by iteratively projecting learnt representations of the classifiers onto the null-space of the target class predictor. RLACE \cite{ravfogel2022linear} models the objective instead as a constrained minimax game. However, recent work shows that  spurious correlations are closely entangled with rest of the sentence representation~\cite{kumar2022probing,he2022controlling}, hence latent space removal methods often unintentionally remove task critical information too, leading to a degradation in model's performance.

\noindent \textbf{Weighting Methods. }
Debiased Focal Loss (DFL) \& Product of Experts (PoE) \cite{mahabadi2019end} are two  methods which leverage a biased model (which relies heavily  on spurious features for prediction) to aid training. Specifically DFL reweighs the samples such that  samples belonging to the majority group are weighed less. PoE models the task as product of two models, where one model is limited in capacity and hence captures the spurious features, where as the other learns non-spurious features. More recent versions can work without annotations for the spurious features~\cite{orgad2022debiasing}, but all methods rely on reweighing the training data. %Maximizing accuracy on the reweighted data may or may not correspond to higher accuracy across majority and minority groups. %weighting removes the need for labeled spurious features. It reformulates its objective to predict main model's success on a sample via an auxilary model and hence upweighs samples with less probability of succeeding (i.e. minority group). 
%But these methods don't explicitly model the feature effect and provide no information on the learnt feature effect. 
% on  deal with features which they assume have 0 effect on the label. They hence implicitly enforces a binary nature on all features for being either causal or spurious. 
%Additionally, they might lead to to a drop in overall accuracy to due new features being learnt which are useful for classification on the re-weighted data and might not be useful for the original target distribution. 

\noindent \textbf{Counterfactual Augmentation.} These methods require collection of counterfactual labeled data that can be used to regularize a classifier~\cite{kaushik2019learning,lu2020gender,gupta2022mitigating}. Obtaining labels for the augmented data is often prohibitively expensive. %These methods require expensive data collection. Furthermore biases might exacerbated by new spurious correlations introduced in the counterfactually augmented data \cite{joshi-he-2022-investigation}.

\paragraph{Comparison to our work.} All above techniques are specific ways to  \textit{remove} the impact of a spurious feature on the classifier. In comparison, we provide a general method that allows us to \textit{control} the learned effect of a spurious feature: one can estimate the effect of a feature on the ground-truth label (which may or may not be zero) and enforce that effect on the classifier. \cite{he2022controlling} make a similar argument against complete removal of spurious features in the context of gender bias and rationale-based methods, while we focus on general spurious correlations and general NLP classifiers.   \cite{joshi2022all}  characterise spurious correlations by necessity and sufficiency and argue for a more finegrained treatment of spurious features.  In terms of implementation, our method can be seen as an extension to the counterfactual augmentation method where we automatically infer the labels for new inputs based on the modified feature's causal effect.

%\paragraph{Spurious Correlation Detection} Prior works uses saliency methods to detect important features for prediction. But differentiating between a important feature being spurious or causal for task requires human intervention. The need for human intervention for spurious feature detection hinders comprehensive, automated and proactive discovery of spurious correlations at scale. 
% Recent work solve this problem for detection of new spurious correlations They take the problem of better attributing input features in dataset \cite{pezeshkpour2021combining} and Recent work analyse feature stability across corpus work to construct methods for distinguishing spurious features from causal features without human intervention \cite{wang2021identifying}. But these methods classifying features as spurious or causal, place an implicit assumption on the binary nature on all features (of being either causal or spurious). This assumption is not often useful in real-world settings (e.g., as we saw in IMDB ratings).

%\paragraph{Controlled bias} Recent work \cite{he2022controlling} also makes an argument parallel to ours and advocates against complete removal of spurious features from the representations. Their work is focused on gender bias and rationale based methods, while our work is for general spurious correlations and general NLP based classifiers. 

\section{Estimating feature effects on labels}
\label{sec:riesz}
%% Intro sub

% \todo{3.1 Background, 3.2 Reisz, Reisz. Problem is denominator will explode, with high predictive. Past work Small derivation }

\begin{figure}[t]
\includegraphics[width=0.4\textwidth]{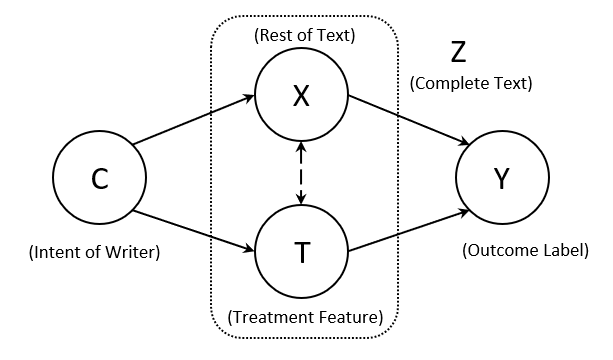}
\caption{Causal Graph for text classification. $C$ is the intent of the writer and a hidden confounding variable. $Z$ is the complete text which is conceptually decomposed as $(X,T)$. $T$ is the treatment feature (the feature of interest) and $X$ is rest of the text. The outcome label $Y$ depends on complete text $Z$.}
\label{fig:cg}
\vspace*{-15pt}
\end{figure}

% \todo{victor. Z and C same. C is whatever}

% Having motivated the need for estimating the need for estimating the effect of feature on the target label, 

Our task is to estimate the effect of text features  on the label $Y$  in training dataset. %for the distribution $\mathcal{D}$. 
This is important for many use cases : \textbf{1)} regularising a text classifier to obey the feature's effect on the label in its prediction; \textbf{2)} identifying annotator artifacts \cite{sap2021annotators} for the label $Y$ in the dataset, e.g., when the estimated effect does not match the ground-truth known effect of a feature. For \textbf{1)}, we present an automated augmentation algorithm in Sec~\ref{sec:cer} based on the estimated feature effect. For \textbf{2)}, we use the feature effect estimation technique and present results on a comment toxicity dataset in Sec~\ref{subsec:eval_annotator}. 

For feature effect estimation, we assume that the data is generated from a distribution $\mathcal{D}$ following the causal graph in Fig.~\ref{fig:cg} \cite{joshi2022all,gui2022causal}. The writer has some intent $C$, which generates the input sentence $(Z)$. The sentence $Z$ can conceptually be disentangled into 2 parts, 1) the feature of interest $(T \in \{0,1\})$ and 2) rest of the text $X$. Annotators perceive the outcome label $(Y)$ from the complete text $Z$. The samples $\{(Z_i,Y_i)\}$ are drawn independently from $\mathcal{D}$. 
Note that the same dataset may contain multiple features $T^j$ ($j=1 ... m$) whose effect needs to be estimated, leading to a different decompositions $(X^j, T^j)$. %\amit{added the above line to say that treatments can vary}.

We term the feature $T$ as \textit{treatment}, and $X$ as \textit{covariates}, following the causality literature. 
Since the variables $X$ and $T$ are sampled from the same latent variable $C$, they are  not independent of each other. For example, in context of IMDB data, if the intent of the writer is to write a positive review then it is highly likely that $X$ will contain positive adjectives while treatment $T$ might be the inclusion of rating as the string \texttt{9/10}. This unobserved latent variable (intent of writer) is called the \textit{confounder} $C$. The correlations between treatment feature $T$ and rest of text $X$ due to the presence of confounder $C$ can lead to the classifier model learning incorrect effect for the treatment feature.  
For computing feature effect, we leverage the causal inference literature~\cite{pearl2009causality, imbens2015causal} and estimate \textit{Average Treatment Effect (ATE)}. 

\subsection{Background}

\paragraph{Definitions.} 

\textit{Propensities}~\cite{pearl2009causality} model the probability of a covariate being treated i.e. $T=1$. They can hence be written as $\mathcal{P}(X)=P(T=1|X)$. %\todo{use a different symbol, curly p for propensity. confusing equation}
\textit{Overlap} is defined as the condition when any covariate $X$ has a non-zero probability of $T=1$ and $T=0$ i.e. $0<P(T|X)<1$ for all $X$. Overlap is a necessary condition for causal effect estimation. 
\textit{Counterfactual : } Given an input $Z=(X,T)$, a counterfactual input is defined as $Z^C=(X,1-T)$, i.e. an input with treatment flipped and rest of the inputs kept constant. The original sample is called the \textit{factual} input. 

\noindent \textbf{Average Treatment Effect (ATE).}
It is defined as the change in label $Y$ on changing treatment $T$ from $0\to1$ keeping everything else constant. %Hence it is mathematically defined as %amit: this line is redundant
\begin{equation*}
\mathbb{E}_X[Y|X,\rmdo(T=1)]-\mathbb{E}_X[Y|X,\rmdo(T=0)]
\end{equation*}
where $\rmdo()$ is the do-operator~\cite{pearl2009causality}, implying an \textit{interventional} change in treatment $T$ while the  covariates $X$ are kept constant. Assume an oracle model  $g_0$ for the task, defined as
$g_0(X,T=t) =\mathbb{E}[Y|X,\rmdo(T=t)] $. Removing the $\rmdo$ notation, $\ate$ estimate can succinctly be written as,  
\begin{equation} \label{eq:ate}
% \hat{\ate} = \frac{1}{n}\underset{i}{\sum}\  (g_0(X_i,1)-g_0(X_i,0))
\ate = \frac{1}{n}\underset{i}{\sum}\  (g_0(X_i,1)-g_0(X_i,0))
% \begin{split}
% \ate = &\mathbb{E}_X[(g_0(X,1)-g_0(X,0))] \\
% \approx &\frac{1}{n}\underset{i}{\sum}\  (g_0(X_i,1)-g_0(X_i,0))
% \end{split}
\end{equation}
The above equation requires access to the oracle model $g_0$ which correctly outputs the label for counterfactual inputs $Z^C$. 

An alternate formulation for computing ATE utilises propensities (of treatment $T$) i.e. $\mathcal{P}_0(X_i)$ instead of the oracle model. The ATE using this formulation is $\mathbb{E}_X[\alpha_0(Z)Y]$ ($\alpha_0$ defined below in Eq~\ref{eq:alpha_def}).
%An alternative estimate of the ATE doesn't require access to the oracle model $g_0$, but  uses
Hence the ATE estimate is
\begin{equation} 
\label{eq:ate_pr}
\ate = \frac{1}{n}\underset{i}{\sum}\  \alpha_0(Z_i) Y_i.
% \hat{\ate} = \frac{1}{n}\underset{i}{\sum}\  \alpha_0(Z_i) Y_i.
\end{equation}
where 
\begin{equation} 
\label{eq:alpha_def}
\alpha_0(Z_i) = (\frac{T_i}{\mathcal{P}_0(X_i)} - \frac{1-T_i}{1-\mathcal{P}_0(X_i)})
\end{equation}
are the \textit{multipliers} computed from propensities.
% We term $\alpha_0$ as $\alpha_\pr$ whenever propensities $p(X_i)$ are used to compute $\alpha_0$.

% \todo{say here that alpha0 is alphaPR}

\paragraph {Direct Estimate.} The simplest method for estimating the average treatment effect is by training a model $g(.)$ as an approximation of the oracle $g_0(.)$ using the loss 
$g = \arg \min_g \mathbb{E}_{\mathcal{D}}[\mathcal{L}(Y,g(Z))]$. The direct estimate of the $\ate$ can then be computed by substituting $g_0(.)$ by $g(.)$ in Eqn.~\ref{eq:ate}. This gives the direct estimate~\cite{shalit2017estimating},  
\begin{equation} \label{eq:ate_direct}
\hat{\ate}_{\dir} = \frac{1}{n}\underset{i}{\sum}\  (g(X_i,1)-g(X_i,0))
\end{equation}
The problem with using the direct estimate is that, in cases where $T$ is correlated with $X$ under $\mathcal{D}$, a loss optimizing method might exploit spurious correlations between $X$ and $T$ to learn a biased model $g(.)$. That is,  the model might over(or under)-estimate the effect of $T$ on the output $Y$. This leads to a biased $\hat{\ate}$.

\paragraph{Propensity-based Doubly Robust (DR) Estimate.}
To resolve the issue of a biased model $g$, DR estimator~\cite{kang2007demystifying,veitch2020adapting} utilises propensities. Since the true propensities $\mathcal{P}_0$ are unknown we learn these propensities using the loss $\mathcal{P}_\pr = \underset{\mathcal{P}}{\mathrm{arg\,min}}\ \mathbb{E}_{\mathcal{D}}[\mathcal{L}(T,\mathcal{P}(X))]$ giving estimated multipliers $\alpha_\pr(Z_i)$. 
% combines equations Eqn~\ref{eq:ate},Eqn~\ref{eq:ate_pr} to give "doubly" robust estimator 
\begin{equation} \label{eqn:ate_dr_pr} \small
\hat{\ate}_{\dr,\pr} = \hat{\ate}_{\dir} + \frac{1}{n}\underset{i}{\sum}\  \alpha_\pr(Z_i)(Y_i-g(Z_i))
\end{equation}
% where $\alpha_\pr(Z_i) = (\frac{T_i}{\mathcal{P}(X_i)} - \frac{1-T_i}{1-\mathcal{P}(X_i)})$ 
% uses estimated propensities to compute $\alpha_0$. 
%Intuitively the "doubly" robust estimate is named so as it's correct if either $g$ is unbiased, or $\alpha_\pr$ is correct.
The DR estimator corrects the bias in $g$ using the correction term (second term in Eqn~\ref{eqn:ate_dr_pr}). If $g$ is systematically wrong on a minority group of examples, their residual error will add up in the correction term. Also, weighing by $\alpha_\pr(Z_i)$ breaks correlation between $X$ and $T$, giving an unbiased correction.  %\todo{correct this prev line}. 
% A doubly robust (DR) estimator~\cite{kang2007demystifying} builds up on the Direct estimate and is asymptotically unbiased. 
% The DR estimator learns a weight for each input sample $Z_i$, denoted by $\alpha(Z_i)$, based on the frequency of $Z_i$ in $\mathcal{D}$. 
%Propensity-based DR estimate uses propensities $p(X)$ ~\cite{veitch2020adapting},  for computing the weighing as .

% Errors in prediction using the biased model $g(.)$ i.e. $(Y_i-g(Z_i))$ are used to compute the correction term. Errors are multiplied by their weighings averaged over the samples to yeild the correction term. Mathematically, the DR estimate is :  \amit{why are error terms introduced?this needs to be explained better. you can give intuition: doubly robust corrects for the bias by looking at the residual of g and then weighs it. }
% where $\alpha_0(Z_i) = (\frac{T_i}{p_0(X_i)} - \frac{1-T_i}{1-p_0(X_i)})$ is the multiplier. 
% \todo{the above paragraph has redundancies. need to shorten it}

\subsection{Riesz Representer (RR) Estimator}
\label{subsec:rr}
While propensity-based methods are the most popular for estimating treatment effect, they suffer from high variance when $P(T=1|X)$ is close to either 1 or 0~\cite{swaminathan2015self}, due to the %. While propensity-based DR is asymptotically unbiased, when estimating the same from finite samples, 
propensity terms in the denominator of the multipliers $\alpha_\pr(.)$. % lead to high variance of the estimate. 
This is especially a problem in high-dimensional text data, where given a treatment $T$ (e.g., a token) the probability of it occurring with most covariate texts $X$ may be close to 0 (e.g., if the covariate $X$ is about a happy incident, probability of a token like "kill" occurring in the sentence is near 0). 
Therefore, we propose a doubly robust estimator for text data based on recent work~\cite{chernozhukov2022riesznet} that avoids estimating the propensities as an intermediate step. Instead it models the coefficient $\alpha_\pr(Z)$ directly.

%The DR estimator requires the multipliers $\alpha(.)$ for the estimation. Modeling propensities is a much harder task towards to goal and hence is sub-optimal. Recent work  proposes modeling these multipliers directly as described below. 

The proposed method depends on the Reisz representation theorem~\cite{chernozhukov2018double}.
% which states that for all bounded functions of Y,  $\mathbb{E}[f(Z)^2] < \infty$, there exists an  $\alpha_\rr(.)$ s.t. 
% %\begin{equation} \small \label{eqn:reisz_org}
% $\mathbb{E}[f(X,1)-f(X,0)] = \mathbb{E}[\alpha_\rr(Z)f(Z)]$.
% %\end{equation}
% %for all $f(.)$. 
% % While $\alpha_0$ can only be used to compute ATE averaging using labels $Y_i$, 

\begin{unnumberedtheorem}[Riesz Representer Theorem]
For a square integrable function $f(Z)$ (i.e. $\mathbb{E}[f^2(Z)] < \infty$), there exists a square integrable function $\alpha_\rr(Z)$ such that 
$$\mathbb{E}[m((Y,Z);f)] = \mathbb{E}[\alpha_\rr(Z)f(Z)]$$
if and only if $\mathbb{E}[m((Y,Z);f)]$ is a continuous linear functional of $f$. 
\end{unnumberedtheorem}

Since the moment functional in ATE formulation (i.e. $m((Y,Z);f) = f(X,1)-f(X,0)$) is indeed a continuous linear functional of $f$, Riesz theorem for our purposes can be written as : 
$$\mathbb{E}[f(X,1)-f(X,0)] = \mathbb{E}[\alpha_\rr(Z)f(Z)]$$ 
for a square integrable function $f$. 
Taking $f$ as $g_0$ (assuming $g_0$ is square integrable), LHS of the equality ($\mathbb{E}[g_0(X,1)-g_0(X,0)]$) is exactly the ATE and the RHS ($\mathbb{E}[\alpha_\rr(Z)g_0(Z)]$)  can be interpreted as a weighted average, as in the propensity formulation of ATE (Eqn.~\ref{eq:ate_pr}). %becomes the ATE estimate form in Eqn~\ref{eq:ate_pr}. 
This means that $\alpha_\rr$ serves as an alternative formulation for $\alpha_0$.
Thus, rather than using the inverse of learnt propensities $\mathcal{P}_\pr$ (i.e. $\alpha_\pr$), we can use the Riesz Representer function $\alpha_\rr$ as an approximation for $\alpha_0$.
% in Eqn.~\ref{eq:ate_pr} and hence also in Eqn.~\ref{eqn:ate_dr_pr}.  

The challenge now remains on how we can estimate the $\alpha_\rr$ function. %For learning $\alpha_\pr$, we had to learn the propensity function. 
To derive an estimation method for $\alpha_\rr$, we use its definition from the Riesz Representation theorem,  i.e.,  $\alpha_\rr(Z)$ weighed by any bounded function $f(Z)$ gives $\mathbb{E}[f(X,1)-f(X,0)]$, as done by \citet{chernozhukov2022riesznet}. 
% This formulation hence gives $\alpha_0$ required in Eqn~\ref{eq:ate_pr} as $\alpha_\rr$. 
% This is used to derive a objective for learning $\alpha_\rr$ as : 
% We can model Riez representer-based $\alpha_\rr(.)$ using loss : 
\begin{equation*}\label{eqn:rr_loss} \normalsize
\begin{split}
\alpha_\rr & = \underset{\alpha}{\mathrm{arg\,min}}\ \mathbb{E}[(\alpha_\rr(Z)-\alpha(Z))^2] \\
& = \underset{\alpha}{\mathrm{arg\,min}}\ \mathbb{E}[\alpha_\rr(Z)^2-2\alpha_\rr(Z)\alpha(Z)+\alpha(Z)^2] \\
& = \underset{\alpha}{\mathrm{arg\,min}}\ \mathbb{E}[-2\alpha_\rr(Z)\alpha(Z)+\alpha(Z)^2] \\
& = \underset{\alpha}{\mathrm{arg\,min}}\ \mathbb{E}[-2(\alpha(X,1)-\alpha(X,0))+\alpha(Z)^2] \\
\end{split}
\end{equation*}
% $$
% \alphahat = \underset{\alpha}{\mathrm{arg\,min}}\ \mathbb{E}[-2 (\alpha(X,1)-\alpha(X,0))+ \alpha(X,T)^2] 
% $$
% \paragraph{Riesz Representer(RR) based DR} 
%\amit{you need to provide more explanation for each step above. What is \alphaRR? is it optimal? you are doing simple optimization, and then how you use the reisz representer theorem.}
% \pari{define more}
The first step is a trivial equality, which says that $\alpha_\rr$ is the solution for the equation $ \underset{\alpha}{\mathrm{arg\,min}}\ \mathbb{E}[(\alpha_\rr(Z)-\alpha(Z))^2]$. %We expand the whole squared term to get the second equation. 
In the third step,  $\alpha_\rr(Z)^2$ can be ignored as the minimization is over $\alpha$ and then we use the Riesz Representation theorem to expand the term $\mathbb{E}[\alpha_\rr(Z)\alpha(Z)]$ as $\mathbb{E}[\alpha(X,1)-\alpha(X,0)]$, thus getting rid of $\alpha_\rr$ and providing an optimization objective. % Eqn~\ref{eqn:rr_loss}, which should give $\alpha_\rr$.

The new learnt riesz function $\alpha_\rr$ can then be used for computing our Doubly Robust estimate. We can simply substitute $\alpha_\pr$ in the DR estimate Eqn~\ref{eqn:ate_dr_pr} by $\alpha_\rr$, giving us RR-based $\hat{\ate}$,
% We can also directly estimate of $\alpha$ without using $\phat$. giving us $\alphahat$   

\begin{equation}\label{eqn:ate_dr_rr} \small
\hat{\ate}_{\dr,\rr} = \hat{\ate}_{\dir} + \frac{1}{n}\sum_i\  \alpha_\rr(Z_i)(Y_i-g(Z_i))
\end{equation}

\begin{table*}
\centering
\resizebox{\textwidth}{!}{%
\begin{tabular}{|l|l|lll|lll|}
\hline
\multirow{2}{*}{$\tau$} & \multirow{2}{*}{Method}  & \multicolumn{3}{c|}{DistilBERT}  & \multicolumn{3}{c|}{BERT} \\
\cline{3-8}
& & 1\% Overlap & 5\% Overlap & 10\% Overlap & 1\% Overlap & 5\% Overlap & 10\% Overlap\\
\hline
\multirow{3}{*}{0.10} & \texttt{Direct} & 15.23 $\pm$ 5.50 & 5.92 $\pm$ 1.31 & 0.48 $\pm$ 1.65 & 8.38 $\pm$ 2.90 & 1.80 $\pm$ 4.66 & 1.13 $\pm$ 0.47 \\
& \propen & \textbf{5.81 $\pm$ 2.76} & 9.80 $\pm$ 1.52 & 6.59 $\pm$ 0.48 & 8.53 $\pm$ 3.77 & 9.83 $\pm$ 5.30 & 6.01 $\pm$ 1.04 \\
& \riesz & \textbf{5.91 $\pm$ 4.35} & \textbf{2.04 $\pm$ 1.25} & \textbf{1.11 $\pm$ 0.62} & \textbf{2.68 $\pm$ 1.24} & 2.61 $\pm$ 0.24 & \textbf{0.88 $\pm$ 0.74} \\
\hline 
\multirow{3}{*}{0.30} & \texttt{Direct} & 18.79 $\pm$ 6.36 & 13.86 $\pm$ 4.64 & 5.94 $\pm$ 0.83 & 22.06 $\pm$ 10.20 & 4.38 $\pm$ 4.77 & 4.72 $\pm$ 5.74 \\
& \propen & 23.48 $\pm$ 2.70 & 20.48 $\pm$ 0.45 & 10.23 $\pm$ 1.19 & 29.02 $\pm$ 5.99 & 23.57 $\pm$ 4.04 & 9.61 $\pm$ 2.79 \\
& \riesz & 16.45 $\pm$ 2.17 & \textbf{0.21 $\pm$ 1.89} & \textbf{1.45 $\pm$ 0.22} & \textbf{0.62 $\pm$ 5.31} & \textbf{2.92 $\pm$ 0.81} &\textbf{2.60 $\pm$ 1.09} \\
\hline
\multirow{3}{*}{0.50} & \texttt{Direct} & 16.95 $\pm$ 3.73 & 11.07 $\pm$ 2.21 & 7.51 $\pm$ 1.56 & 20.36 $\pm$ 1.44 & 17.42 $\pm$ 1.62 & 11.59 $\pm$ 2.45 \\
& \propen & 61.88 $\pm$ 11.10 & 36.11 $\pm$ 2.73 & 17.09 $\pm$ 1.41 & 47.28 $\pm$ 11.27 & 31.41 $\pm$ 5.72 & 13.16 $\pm$ 4.02 \\
& \riesz & 15.62 $\pm$ 3.28 & \textbf{1.50 $\pm$ 1.39} & \textbf{2.73 $\pm$ 0.28} & \textbf{1.42 $\pm$ 3.37} & \textbf{1.53 $\pm$ 1.62} & \textbf{0.11 $\pm$ 0.91} \\
\hline
\end{tabular}
}
\caption{\label{tab:reiszpropen_est}
MAE (x100) of feature effect estimate. \riesz gives lower error MAE error than \stdfine, across values of overlap and true feature effect. %Riesz gives gains in MAE in low overlap settings and high ATE settings. 
\propen shows high MAE error (especially in lower overlap setting). }
\vspace*{-5pt}
\end{table*}

\section{Controlling learnt effects in a classifier}
\label{sec:cer}
% While there labeling of counterfactual data given an effect (this section)
% \todo{prev section ends when able to write causal effect. give an algorithm using the causal effect. Section 4 for peeps only interested in ML. }
% While there labeling of counterfactual data given an effect (this section)

%Here we discuss method for solving this issue.
Armed with an estimator of feature effect on the label, we now describe methods to enforce the feature effect on  a predictive model's output. Given data $\{(Z,Y)\}$ where $Z$ are input sentences and $Y$ is output label, the goal is to learn a predictive model $f$ for $Y$ such that the causal effect of a feature on $f(Z)$ is the same as the true feature effect, $\tau^j$ for the $j$th feature. 
That is, $\tau^j$ should be equal to $\mathbb{E}_{\mathcal{D}} [f(X^j, T^j=1) - f(X^j, T^j=0)]$ where $X^j$ refers to all input features except $T^j$ and the expectation is over the training distribution. %\todo{change this to Xj Tj notation from eqn 9}
As discussed in Section~\ref{sec:riesz}, the ideal predictive function is $g_0$ since it will ensure the correct feature effect,$\tau^j = \mathbb{E}_{\mathcal{D}} [g_0(X^j, T^j=1) - g_0(X^j, T^j=0)]$, and will also provide high accuracy since it is the true data generating function. 

%That In other cases, we may not require enforcing exactly $\tau^j$, but 
\subsection{Counterfactual-based Regularisation}
\label{subsec:cer_reg}
% To resolve this prior work takes multiple approaches, 1) assumes treatment effect to be 0 and imposes invariance constraints, 2) upweighs certain samples while downweighing others 3) collects counterfactual labeled data \cite{kaushik2019learning}.

% Points : 1) there are no guarantees that the methods will give the desired effect
% 2) might lead to unwanted correlations on upweighing samples
% 3) single loss is best

%\amit{the above two paragraphs and this line should go in related work. Here we should start directly by describing the method.}
To approximate the oracle function $g_0(Z)$, for a given loss $\mathcal{L}$, Standard ERM loss minimisation optimizes, 
%\begin{equation*}
 $   \arg \min_f {\mathbb{E}_{\mathcal{D}}}[\mathcal{L}(Y,f(Z))]$. 
%\end{equation*}
But machine learning data is often \textit{underspecified} \cite{d2020underspecification,lee2022diversify}, leading to the ERM returning multiple solutions $f$ with similar accuracy on validation set. 
These different solution $f$ weigh different features in input text differently. As a result, the obtained solution can be far from $g_0$. %hence behave very differently on out-of-distribution data \pari{want to mention OOD? or even the complete line?}. 

Therefore, we use the provided feature effect to constraint the solution space. A first idea is to add a regularization term that aligns the model's learnt feature effect with the provided effect.
Suppose that we are given a list of $m$ binary features $\{T^j\}_{1\ldots m}$ which are suspected to have a spurious correlation (e.g., such features can be discovered using explanation methods on an ERM model~\cite{wang2021identifying}). We can conceptually decompose an input sentence $Z$ into $m$ different pairs $\{(X^j,T^j)\}_{1\ldots m}$, where $X^j$ is the part of the sentence $Z$ apart from $T^j$. Then using the given feature effect $\{\tau^j\}_{1\ldots m}$ for each feature, we can write the regularized loss,
\begin{equation}\label{eqn:cerate_reg}
    \mathcal{L} + \lambda\frac{1}{m}\sum_j (f(X^j,1)-f(X^j,0)-\tau^j)^2
\end{equation}
where $\lambda$ is the regularisation constant. 

While we proposed regularizing to $\tau^j$, sometimes one may want to completely remove a feature's effect based on domain knowledge. For example, a biased dataset may exhibit a non-zero feature's effect on the label, but due to fairness reasons, one would like to completely remove its effect. In that case, we can simply set $\tau^j=0$ and apply Equation~\ref{eqn:cerate_reg}. When $\tau^j$ is set to zero, \cero can be seen as optimizing the same objective as methods that aim to fully remove the feature's effect~\cite{ravfogel2020null,mahabadi2019end}.

\subsection{Augmentations for Estimated Effect}
\label{subsec:cer_aug}
%\amit{before describing augmentations, you need to define the regularization objective. that is more intuitive. }
We also consider a data augmentation alternative to regularization. 
Given distribution $(Z,Y) \sim \mathcal{D}$,  $m$ binary features $\{T^j\}_{1\ldots m}$, and  their feature effects $\{\tau^j\}_{1\ldots m}$, we can augment along any of the $m$ features to generate a counterfactual distribution. 
When we augment along the $j$ feature, the new input becomes $Z^{j,C} = (X^{j},1-T^{j})$. Using the feature's effect $\tau^j$, we can estimate the corresponding label $Y^{j,C}$ for the input $Z^{j,C}$. Intuitively, a higher feature effect makes it more likely that the label will change (see Supp~\ref{sec:app_label_flip} for details). We get a new counterfactual distribution,  $(Z^{j,C},Y^{j,C}) \sim \mathcal{D}^{j,C}$.
% \todo{seems vague, need to give some intuition on how label changes. also, is it flipped? what about regression? }

Similarly other counterfactual distributions can be found, giving us $\{\mathcal{D}^{j,C}\}_{1\ldots m}$. A union can be taken over these distributions to give us the counterfactual distribution over these $m$ features as $\mathcal{D}^C = \cup_{j=1}^m \mathcal{D}^{j,C}$
This new generated distribution can then be included in training as counterfactual augmentations while minimising the loss, 
% $$
% \fhat = \underset{f}{\mathrm{arg\,min}}\ {\mathbb{E}_{\mathcal{D}+\lambda \mathcal{D}^C}}[\mathcal{L}(Y,f(Z))]
% $$
\begin{equation}\label{eqn:cerate_aug} \small
    \arg \min_f {\mathbb{E}_{\mathcal{D}}}[\mathcal{L}(Y,f(Z))]+\lambda \mathbb{E}_{\mathcal{D}^C}[\mathcal{L}(Y,f(Z))]
\end{equation}
where we now draw samples from the combined distribution $\mathcal{D}+ \mathcal{D}^C$. $\lambda$ signifies the weighting of samples drawn from augmented counterfactual distribution $\mathcal{D}^C$ in the loss function. 

While both regularisation and data augmentation can help us control the learned effect of features,   owing to the scalability and ease of optimization,  we use the augmentation version of our algorithm to present our results.

\subsection{FEAG: Two-phase algorithm}
\label{subsec:cer_algo}
 To summarize, the proposed algorithm,  Feature Effect Augmentation (\cero), proceeds in two phases. 
 It takes as input a set of features $T^j: j=1 ... m$,  that may be suspected to be spurious, which can be derived using an automated saliency method (e.g., top-k important tokens)~\cite{pezeshkpour-etal-2022-combining,wang2021identifying} or  based on domain knowledge.

\noindent \textbf{Feature effect estimation.} %\todo{need some justification for 2-headed model. need to cite dragonnet intuition and redraw architecture in suppl.}
%We assume we are given a set of features $\{T^j\}_{1\ldots m}$. These features can be from an automated saliency method \cite{pezeshkpour-etal-2022-combining} or from a moderator. For automated saliency detection, these features are taken as the list of top-K most important tokens that affect a model's decision (as given by saliency methods like input-gradients or occlusion-based) }.
% \todo{Here can you write 1-2 lines on how one may obtain such features. you can write similar to diyi yangs emnlp findings paper--also cite them}
For each of the features $T^j$, we estimate the feature effect using the Reisz estimator from Section 3.2. We follow the  2-headed model architecture with shared parameters \cite{shi2019adapting} to learn the Riesz representer $\alpha_\rr$ and the model $g$ for $Y$ (details are in Supp~\ref{subsec:app_twohead}, Fig~\ref{fig:riesz}). 
% \todo{move the commented text to new section in appendix: architecture of feature effect estimation and link it here}
Note that $\alpha_\rr$ and $g$ should share sentence representation extraction module to ease learning~\cite{chernozhukov2022riesznet} (i.e., they have the same BERT model, but different final layer linear heads). 
These learnt models can be used in Eqn~\ref{eqn:ate_dr_rr} to get feature effect estimates ($\{\tau^j\}_{1\ldots m}$) on held-out data. %We can then use the feature effect estimates  for the next step.

\noindent \textbf{Counterfactual Augmentation.}
Our modular pipeline allows practitioners to change the feature estimate $\tau^j$ according to their needs before using them for counterfactual augmentations. 
Using the features and their effect estimates, we create  counterfactually augmented data $\mathcal{D}^C$ as described in Sec~\ref{subsec:cer_aug} and include them  while training (Eqn~\ref{eqn:cerate_aug}) to learn the final classifier.

\begin{table}
\centering
\resizebox{0.49\textwidth}{!}{%
\begin{tabular}{|l||ll|ll|}
\hline
% \multirow{2}{*}{Method} & \multicolumn{3}{c|}{BERT} & \multicolumn{3}{c|}{DistilBERT} \\
% & CC Random SS & CC Kill & IMDB  
% & CC Random SS & CC Kill & IMDB  
% \\
% \hline
% \stdfine & 59.35 $\pm$ 3.52 & 36.91 $\pm$ 1.31 & 71.93 $\pm$ 9.36 & 56.71 $\pm$ 4.72 & 35.15 $\pm$ 2.14 & 66.42 $\pm$ 9.12 \\
% \riesz & 54.57 $\pm$ 0.37 & 32.13 $\pm$ 1.06 & 52.51 $\pm$ 2.63 & 55.38 $\pm$ 0.05 & 33.84 $\pm$ 0.80 & 55.37 $\pm$ 0.77 \\
\multirow{2}{*}{Method} & \multicolumn{2}{c|}{BERT} & \multicolumn{2}{c|}{DistilBERT} \\
\cline{2-5}
& CC Sub. & IMDB 
& CC Sub. & IMDB 
\\
\hline
\stdfine & 18.46 $\pm$ 0.61 & 71.93 $\pm$ 9.36 & 19.07 $\pm$ 0.67 & 66.42 $\pm$ 9.12 \\
\riesz & 15.77 $\pm$ 0.50 & 52.51 $\pm$ 2.63 & 15.14 $\pm$ 0.63 & 55.37 $\pm$ 0.77 \\
\propen & 36.25 $\pm$ 4.88 & 45.08 $\pm$ 10.05 & 24.20 $\pm$ 0.98 & 56.86 $\pm$ 6.75\\
\hline
\end{tabular}
}
\caption{\label{tab:riez_est}
\riesz estimated feature effect is less than the \stdfine learned feature effect, indicating that \stdfine method over-weighs the treatment feature.}
\vspace*{-10pt}
\end{table}

\begin{table*}
\centering
\footnotesize
\resizebox{\textwidth}{!}{%

\begin{tabular}{|p{0.04\linewidth}|p{0.71\linewidth}|p{0.05\linewidth}|p{0.20\linewidth}|}
\hline
S.No. & Sentence & Riesz & Interpretation  \\
\hline
\multirow{2}{*}{1.} & maybe it's just burning a bunch of islamist terrorists \textbf{killed} in combat & 1.69 & $T = 1 \ \&\  P(T=1|X) \uparrow$\\
& only arabs doing what they do best, \textbf{killing} other arabs. black september, darfur, isis/isil. & 1.87  & $T = 1 \ \&\  P(T=1|X) \uparrow$\\
\hline
\multirow{2}{*}{2.} & "strong stated desire to \textbf{kill} people in the name of islamic state" that is the important part. & 6.13  & $T = 1 \ \&\  P(T=1|X) \downarrow$ \\
& who do you think is \textbf{killing} the women its the male nativees and we hear about this & 6.14  & $T = 1 \ \&\  P(T=1|X) \downarrow$\\
% who do you think is \textbf{killing} the women its the male nativees and we gotta hear about this morning noon and night....... & 6.14  & $T = 1 \ \&\  P(T=1|X) \downarrow$\\
\hline
\multirow{2}{*}{3.} & they also never tell you how often the officer doing the shooting...is black & -3.75  & $T = 0 \ \&\  P(T=1|X) \uparrow$\\
& driving into crowds of people is a popular approach for muslim terrorists & -4.53  & $T = 0 \ \&\  P(T=1|X) \uparrow$\\
\hline
\multirow{2}{*}{4.} & i am getting very tired about hearing anything from this neurotic woman. give it a rest. & -0.56  & $T = 0 \ \&\  P(T=1|X) \downarrow$ \\
& and these men give so much to charity. there is no record of trump's charity contributions. & -0.82   & $T = 0 \ \&\  P(T=1|X) \downarrow$ \\
\hline

\end{tabular}
}
\caption{\label{tab:qualitative_egs}
Examples on \texttt{Kill} keyword in CivilComments Subsampled dataset. %Treated sentences have positive Riesz score. 
Sentences having violent words (other than \texttt{kill}) are assigned a lower score, while sentences having non-violent context are assigned high score.} 
\vspace*{-5pt}
\end{table*}

\section{Experiments}
\label{sec:experiments}
We have three goals for evaluation: \textbf{1)} RR-based estimators of feature effect are more accurate than propensity-based estimators; \textbf{2)} \cero using RR-based estimators provides better overall accuracy while minimizing spurious correlation compared to existing baselines for removing spurious correlations; \textbf{3)} Our feature effect estimator is a general method and can be used to detect annotator bias. 

% \subsection{Experimental Setup}

% \paragraph{Counterfactual generation} 

% \paragraph{Models} We show results on two BERT-based models (BERT and DistilBERT) 

\subsection{Datasets}
\label{subsec:eval_datasets}
Since the true feature effect is unknown for real-world data, we construct a semi-synthetic dataset based on the CiviComments dataset~\cite{borkan2019nuanced}. In addition, we evaluate on subsampled versions of the CivilComments and IMDB dataset.

\noindent \textbf{CivilComments Semi-Synthetic (SS).} 
CivilComments is a toxicity detection dataset $\{(X,Y)\}$, where $X$ are input sentences and $Y$ is the toxicity label (1 means \textit{toxic}).  To evaluate our methods, we need to construct a dataset generated from the causal graph in Fig.~\ref{fig:cg}. 
Since the  writer's intent (confounder) is unknown, we construct it as a property of the input text, $W=h(X) \in \{0,1\}$, leading to the modified causal graph in Fig.~\ref{fig:cg23} (Supp~\ref{sec:alternate_causal_graphs}). %\todo{do not use P for property since it is used for Prob. Use S or anything else.}  % we would ideally like to have a confounder as the intent of the write which would serve as a common cause for both our treatment and covariates. But since the intent of writer is rarely known in real world datasets, we consider an alternate causal graph for our semi-synthetic setup, given in Fig.~\ref{fig:cg2}. Here the confounder instead of being the intent of the writer, is a property of the covariate text, 
To obtain $h(X)$, we train a binary classifier using a DistilBERT model on $(X,Y)$ pairs. % for a few hundred iterations
Finally we sample a new label as $Y' \sim \textrm{Bernoulli}((1-\tau)Y + \tau T)$, giving the true feature effect as $\tau$. %The final graph can be seen in Fig~\ref{fig:cg3}. 
The complete text $Z=(X,T)$ is constructed by prepending each covariate sentence $X$ with the word \treated if $T=1$ and \untreated if $T=0$. 
% \amit{add the resultant graph to appendix and reference it here}

\begin{table*}
\centering
\resizebox{\textwidth}{!}{%
\begin{tabular}{|l||llll|HHllH|}
\hline
Method & Group1 & Group2 & Group3 & Group4 & Treatment0 & Treatment1 & Total & Avg Group & Direct \\
\hline
\stdfine & 99.46 $\pm$ 0.08  & 3.52 $\pm$ 0.80  & 1.61 $\pm$ 0.29  & 99.42 $\pm$ 0.10  & 88.30 $\pm$ 0.03  & 87.25 $\pm$ 0.02  & \textbf{87.77 $\pm$ 0.02}  & 51.00 $\pm$ 0.17  & 59.43 $\pm$ 1.28  \\
\removetoken & 88.71 $\pm$ 0.75  & 28.06 $\pm$ 0.94  & 37.46 $\pm$ 2.36  & 90.69 $\pm$ 0.85  & 82.86 $\pm$ 0.46  & 82.74 $\pm$ 0.62  & 82.80 $\pm$ 0.14  & 61.23 $\pm$ 0.45  & 1.43 $\pm$ 0.13  \\
\hline
\dfl & 72.45 $\pm$ 1.33  & 35.62 $\pm$ 5.51  & 53.58 $\pm$ 2.61  & 82.46 $\pm$ 3.38  & 70.30 $\pm$ 0.95  & 76.51 $\pm$ 2.25  & 73.45 $\pm$ 0.76  & 61.03 $\pm$ 0.77  & 3.78 $\pm$ 1.53  \\
\dflsuccess & 99.22 $\pm$ 0.34  & 4.13 $\pm$ 1.21  & 3.12 $\pm$ 0.92  & 99.34 $\pm$ 0.18  & 88.26 $\pm$ 0.21  & 87.25 $\pm$ 0.01  & \textbf{87.75 $\pm$ 0.10}  & 51.45 $\pm$ 0.41  & 56.99 $\pm$ 2.67  \\
\poe & 100.00 $\pm$ 0.00  & 0.18 $\pm$ 0.14  & 0.00 $\pm$ 0.00  & 99.96 $\pm$ 0.02  & 88.59 $\pm$ 0.00  & 87.30 $\pm$ 0.02  & \textbf{87.94 $\pm$ 0.01}  & 50.03 $\pm$ 0.03  & 74.95 $\pm$ 2.60  \\
\hline
\inlp & 79.10 $\pm$ 3.75  & 73.44 $\pm$ 7.52  & 38.77 $\pm$ 7.53  & 36.35 $\pm$ 9.45  & 74.50 $\pm$ 2.50  & 41.06 $\pm$ 7.31  & 57.54 $\pm$ 2.48  & 56.92 $\pm$ 1.41  & 59.62 $\pm$ 3.91  \\
\hline
\subsample & 85.45 $\pm$ 3.98  & 59.89 $\pm$ 8.49  & 27.59 $\pm$ 8.76  & 57.72 $\pm$ 9.77  & 78.85 $\pm$ 2.53  & 57.99 $\pm$ 7.46  & 68.27 $\pm$ 2.54  & 57.66 $\pm$ 1.55  & 3.25 $\pm$ 1.35  \\
\groupdro & 63.98 $\pm$ 4.43  & 43.18 $\pm$ 4.68  & 59.42 $\pm$ 4.75  & 72.19 $\pm$ 3.31  & 63.46 $\pm$ 3.38  & 68.51 $\pm$ 2.30  & 66.02 $\pm$ 0.97  & \textbf{59.69 $\pm$ 0.28 } & 7.91 $\pm$ 0.81  \\
\hline
\cer{0} & 98.89 $\pm$ 0.48  & 7.48 $\pm$ 1.77  & 4.03 $\pm$ 1.53  & 97.40 $\pm$ 0.76  & 88.07 $\pm$ 0.29  & 85.99 $\pm$ 0.44  & 87.01 $\pm$ 0.34  & 51.95 $\pm$ 0.31  & 47.04 $\pm$ 3.88  \\
\cer{ate} & 98.30 $\pm$ 0.30  & 4.13 $\pm$ 0.94  & 7.75 $\pm$ 1.28  & 99.36 $\pm$ 0.18  & 87.97 $\pm$ 0.14  & 87.27 $\pm$ 0.06  & \textbf{87.62 $\pm$ 0.06}  & 52.39 $\pm$ 0.16  & 51.59 $\pm$ 0.18  \\
\hline
\end{tabular}
}
\caption{\label{tab:ceracc_ss_bert}
Accuracy across groups for CivilComments Semi-Synthetic (0.50 ATE,5\% Overlap), trained using BERT.} % All models are trained using BERT model.  \cer{ate} gives the best tradeoff between total and average  group accuracy.}
\vspace*{-5pt}
\end{table*}

\noindent \textbf{CivilComments Subsampled.} %\amit{subsampled is a better name than Kill. consider renaming dataset.}
Rather than introducing a new treatment, here we subsample CivilComments to introduce a spurious correlation between an existing token \texttt{kill} and label $Y$. Here all sentences with token \texttt{kill} are considered as treated, while others untreated. To exacerbate the spurious correlation between $T$ and $Y$, we subsample our data based on the learnt property $W$ (from above), following the causal graph in Fig~\ref{fig:cg2}. % instead of sampling based on $Y$ we use the learnt property $W$ (from Sec.~\ref{subsec:eval_reisz}) for subsampling. 

\noindent \textbf{IMDB.} 
From the IMDB reviews dataset \cite{maas-EtAl:2011:ACL-HLT2011}, we consider reviews that contain a numerical rating---text string from either the set \{\texttt{7/},\texttt{8/},\texttt{9/}\} or  \{\texttt{2/},\texttt{3/},\texttt{4/}\}. To construct a binary treatment variable, occurrences of these strings are replaced by \treated if the rating is 7, 8, or 9 and an empty string otherwise. %from the dataset, and the sentences originally having \{\texttt{7/},\texttt{8/},\texttt{9/}\} in them are considered treated. 
%The dataset is then subsampled to have equal number of positive and negative sentiment reviews.
The \treated token is predictive of the sentiment with 90\% accuracy. %The test set is constructed similarly. 

For dataset and training details,  see Supp~\ref{sec:app_dataset}, Supp~\ref{sec:app_training} respectively. All results are run for 3 seeds.

\subsection{Evaluating Feature Effect Estimation}
\label{subsec:eval_reisz}
We evaluate the performance of different estimators in Sec~\ref{sec:riesz} on the CivilComments SS dataset (with different overlap $\epsilon$ and feature effects $\tau$). %We also present numbers on IMDB and CivilComments Kill and Qualitative Results on the later.
% \paragraph{Causal Graph} 
% Consider the graph in Fig.~\ref{fig:cg}. We need to know the intent of writer to sample the treatment Since the effect of feature in real world datasets is rarely known, we build a semi-synthetic dataset with known feature effects for the same. 
We compare the \riesz-based DR  estimator (Eqn~\ref{eqn:ate_dr_rr}) with the \stdfine (Eqn~\ref{eq:ate_direct}) and \propen-based DR (Eqn~\ref{eqn:ate_dr_pr}) baselines. All estimators are finetuned using either BERT or DistilBERT as base model.
See Supp~\ref{s}

% \amit{need a Effect Estimator paragraph header--short para, just listing the three algorithms and referencing the corresponding eqn from Sec 3}

\noindent \textbf{Quantitative Results.} Table~\ref{tab:reiszpropen_est} shows the mean error in estimating feature effect across   $\tau \!\in \!\{0.10,0.30,0.50\}$ and $\epsilon\!\in\!\{0.01,0.05,0.10\}$. % For estimating the effect we train a BERT and DistilBERT based classification model $g(.)$. We report the MAE error of the estimate (x100). 
For hyperparameter selection, see Supp.~\ref{sec:app_bestpropensityeval}. 
Across all settings (barring 1\% overlap with high $\tau$), \riesz is able to estimate the effect with low error. \stdfine fails to do well in high $\tau$ and low $\epsilon$ ranges, failing for both $\tau\!=\!0.50$ and $\epsilon\!=\!0.01$. Due to its high variance, \propen is unable to work well, often producing an estimate worse than \stdfine. 

For the two real-world datasets, true feature effect is unknown. But comparing the effect estimates of \stdfine and \riesz, \stdfine tends to over-estimate the feature effect (due to spurious correlation), which is corrected to a lower value by \riesz. % we present the the model's learnt effect (i.e. Direct estimate) and Riesz DR estimates in Tab.~\ref{tab:riez_est}. We can see that for both IMDB and CC Kill the \stdfine model is overweighing the effect of feature $T$. 
% \amit{the below paragraph should go to 5.1, since this is about quality of RR effect estimation}

\begin{table*}[th]
\centering
\resizebox{\textwidth}{!}{%
% \begin{tabular}{|l|lllllllll|}
\begin{tabular}{|l||llll|HHllH|}
\hline
% \multirow{2}{*}{Method} & \multicolumn{9}{c|}{BERT} \\
Method & Group1 & Group2 & Group3 & Group4 & Treatment0 & Treatment1 & Total & Avg Group & Direct \\  
\hline
\stdfine & 76.72 $\pm$ 0.82  & 5.80 $\pm$ 1.57  & 81.72 $\pm$ 0.91  & 96.72 $\pm$ 0.35  & 79.16 $\pm$ 0.36  & 79.92 $\pm$ 0.19  & \textbf{79.38 $\pm$ 0.29}  & 65.24 $\pm$ 0.31  & 18.46 $\pm$ 0.35  \\
\removetoken & 75.63 $\pm$ 0.79  & 15.22 $\pm$ 1.02  & 83.10 $\pm$ 0.43  & 90.15 $\pm$ 0.61  & 79.27 $\pm$ 0.33  & 76.31 $\pm$ 0.33  & 78.40 $\pm$ 0.23  & 66.02 $\pm$ 0.28  & 10.20 $\pm$ 0.09  \\
\hline
\dfl & 83.28 $\pm$ 0.16  & 9.42 $\pm$ 0.59  & 67.82 $\pm$ 0.66  & 94.09 $\pm$ 0.80  & 75.74 $\pm$ 0.33  & 78.45 $\pm$ 0.61  & 76.54 $\pm$ 0.36  & 63.65 $\pm$ 0.24  & 17.53 $\pm$ 0.78  \\
\dflsuccess & 78.80 $\pm$ 1.84  & 3.62 $\pm$ 1.18  & 77.82 $\pm$ 2.34  & 97.54 $\pm$ 0.46  & 78.32 $\pm$ 0.21  & 80.19 $\pm$ 0.22  & 78.87 $\pm$ 0.21  & 64.44 $\pm$ 0.20  & 19.52 $\pm$ 0.52  \\
\poe & 79.02 $\pm$ 0.62  & 10.14 $\pm$ 1.57  & 79.43 $\pm$ 0.66  & 95.24 $\pm$ 0.71  & 79.22 $\pm$ 0.57  & 79.52 $\pm$ 0.33  & \textbf{79.30 $\pm$ 0.37}  & 65.96 $\pm$ 0.52  & 17.57 $\pm$ 1.06  \\
\hline
\inlp & 69.02 $\pm$ 1.04  & 6.52 $\pm$ 2.51  & 88.45 $\pm$ 0.10  & 95.07 $\pm$ 0.57  & 78.49 $\pm$ 0.49  & 78.71 $\pm$ 0.00  & 78.55 $\pm$ 0.34  & 64.77 $\pm$ 0.25  & 18.88 $\pm$ 0.10  \\
\hline
\subsample & 73.99 $\pm$ 0.32  & 28.26 $\pm$ 2.72  & 83.45 $\pm$ 1.14  & 84.40 $\pm$ 0.97  & 78.60 $\pm$ 0.51  & 74.03 $\pm$ 0.29  & 77.25 $\pm$ 0.45  & \textbf{67.52 $\pm$ 0.17}  & 1.80 $\pm$ 0.14  \\
\groupdro & 78.14 $\pm$ 1.32  & 44.93 $\pm$ 4.27  & 73.45 $\pm$ 5.25  & 71.92 $\pm$ 2.36  & 75.85 $\pm$ 2.03  & 66.93 $\pm$ 1.26  & 73.22 $\pm$ 1.79  & \textbf{67.11 $\pm$ 1.20}  & 0.11 $\pm$ 1.65  \\
\hline
\cer{0} & 78.25 $\pm$ 0.45  & 11.59 $\pm$ 1.18  & 79.43 $\pm$ 0.25  & 94.25 $\pm$ 0.35  & 78.82 $\pm$ 0.16  & 78.98 $\pm$ 0.39  & 78.87 $\pm$ 0.14  & 65.88 $\pm$ 0.28  & 9.25 $\pm$ 0.42  \\
\cer{ate} & 78.80 $\pm$ 0.32  & 10.14 $\pm$ 0.59  & 80.34 $\pm$ 0.32  & 95.73 $\pm$ 0.35  & 79.55 $\pm$ 0.30  & 79.92 $\pm$ 0.19  & \textbf{79.66 $\pm$ 0.17}  & 66.25 $\pm$ 0.22  & 14.92 $\pm$ 0.19  \\
\hline
\end{tabular}
}
\caption{\label{tab:ceracc_kill_bert}
Accuracy across groups for CivilComments Subsampled trained using BERT model.}%Last two column show total and average group accuracies. See Sec~\ref{subsec:cer_results} for details}
\vspace*{-5pt}
\end{table*}

\begin{table*}
\centering
\resizebox{\textwidth}{!}{%
\begin{tabular}{|l||llll|HHllH|}
\hline
Method & Group1 & Group2 & Group3 & Group4 & Treatment0 & Treatment1 & Total & Avg Group & Direct \\  
\hline
\stdfine & 98.53 $\pm$ 0.73  & 5.82 $\pm$ 2.16  & 20.78 $\pm$ 8.84  & 99.87 $\pm$ 0.05  & 88.52 $\pm$ 0.57  & 89.47 $\pm$ 0.22  & 88.98 $\pm$ 0.38  & 56.25 $\pm$ 2.25  & 71.93 $\pm$ 5.41  \\
\removetoken & 81.96 $\pm$ 1.69  & 79.37 $\pm$ 1.98  & 69.26 $\pm$ 1.77  & 76.73 $\pm$ 2.67  & 80.32 $\pm$ 1.30  & 77.02 $\pm$ 2.17  & 78.71 $\pm$ 0.82  & \textbf{76.83 $\pm$ 0.50 } & 0.46 $\pm$ 0.09  \\
\hline
\dfl & 96.87 $\pm$ 1.27  & 8.99 $\pm$ 6.72  & 30.30 $\pm$ 9.52  & 99.28 $\pm$ 0.51  & 88.29 $\pm$ 0.28  & 89.30 $\pm$ 0.30  & 88.78 $\pm$ 0.29  & 58.86 $\pm$ 3.00  & 70.05 $\pm$ 3.84  \\
\dflsuccess & 94.82 $\pm$ 0.94  & 7.41 $\pm$ 3.54  & 41.56 $\pm$ 5.34  & 99.67 $\pm$ 0.27  & 87.96 $\pm$ 0.16  & 89.47 $\pm$ 0.17  & 88.70 $\pm$ 0.00  & 60.86 $\pm$ 1.71  & 63.59 $\pm$ 4.03  \\
\poe & 98.59 $\pm$ 0.84  & 14.29 $\pm$ 8.51  & 24.68 $\pm$ 4.25  & 98.82 $\pm$ 0.97  & 89.07 $\pm$ 0.18  & 89.47 $\pm$ 0.14  & \textbf{89.27 $\pm$ 0.16}  & 59.09 $\pm$ 1.51  & 69.40 $\pm$ 3.60  \\
\hline
\inlp & 68.33 $\pm$ 4.57  & 58.73 $\pm$ 14.62  & 49.78 $\pm$ 6.50  & 50.43 $\pm$ 14.88  & 65.94 $\pm$ 3.87  & 51.35 $\pm$ 11.72  & 58.82 $\pm$ 5.45  & 56.82 $\pm$ 1.34  & 68.26 $\pm$ 3.99  \\
\hline
\subsample & 71.53 $\pm$ 3.64  & 65.08 $\pm$ 1.98  & 74.46 $\pm$ 2.90  & 85.67 $\pm$ 2.94  & 71.91 $\pm$ 2.80  & 83.39 $\pm$ 2.40  & 77.51 $\pm$ 0.28  & 74.18 $\pm$ 0.09  & 2.86 $\pm$ 0.21  \\
\groupdro & 79.40 $\pm$ 3.67  & 55.56 $\pm$ 2.70  & 67.97 $\pm$ 1.97  & 90.66 $\pm$ 0.82  & 77.93 $\pm$ 2.95  & 86.78 $\pm$ 0.50  & 82.25 $\pm$ 1.34  & 73.40 $\pm$ 0.51  & 17.07 $\pm$ 2.56  \\
\hline
\cer{0} & 94.63 $\pm$ 0.72  & 33.33 $\pm$ 7.23  & 46.75 $\pm$ 1.84  & 97.30 $\pm$ 1.09  & 88.46 $\pm$ 0.47  & 90.23 $\pm$ 0.24  & \textbf{89.33 $\pm$ 0.15}  & 68.00 $\pm$ 1.65  & 44.84 $\pm$ 2.76  \\
\cer{ate} & 95.46 $\pm$ 1.27  & 15.34 $\pm$ 3.03  & 43.29 $\pm$ 5.49  & 99.34 $\pm$ 0.28  & 88.74 $\pm$ 0.43  & 90.06 $\pm$ 0.13  & \textbf{89.38 $\pm$ 0.16}  & 63.36 $\pm$ 1.75  & 57.26 $\pm$ 2.88  \\
\hline
\end{tabular}
}
\caption{\label{tab:ceracc_imdb_bert}
IMDB dataset; models trained using BERT. \cer{ate} and \cer{0} achieve highest average group accuracy.}
\vspace{-1em}
\end{table*}

\noindent \textbf{Qualitative Results.}
To understand how the Reisz estimator works, we show qualitative results for Civil Comments Subsampled dataset in Table~\ref{tab:qualitative_egs}. To counter the spurious correlation of token \texttt{kill} (T) with other parts of text (X) that cause toxicity (Y), the Riesz estimator provides a low weight to sentences having features X that commonly occur with T, and higher weight to sentences having X that rarely occur with T. Treated samples (T=1) have a positive Riesz value and vice versa. %Also a higher magnitude of Riesz estimator implies that the combination of $(X,T)$ is rarely seen in the input (i.e. it belongs to a minority group) while lower magnitude implies majority. 
We can see that sentences with violent language (in addition to \texttt{kill})  are assigned a low score while other sentences with \texttt{kill} are assigned a high score, thus serving to extract the \textit{isolated} feature effect of \texttt{kill} (without confounding due to other tokens). 
%We can see in the first 2 examples in the table, are about violent topics i.e. have violent covariates $X$ and contain the word "kill" i.e. $T=1$. Hence they are assigned a low Riesz value. Similarly other rows can be analysed. 
%This qualitatively shows that Riesz estimator is accurately able to differentiate between the majority and minority groups. 

%Given these results, we use \riesz for effect estimation for the rest of the analysis.

\subsection{Accuracy of \cero classifiers}

\label{subsec:cer_results}
% \input{tables/cer_accuracies_combined_bert}

% \amit{it may be better to have a single datasets section before 5.1--the current datasets description feels scattered.}
We now compare \cero classifiers based on \riesz, \cer{ate},  and based on zero effect, \cer{0}, with prior debiasing algorithms. % our results the three datasets (Sec.~\ref{subsec:eval_datasets}) constructed on sentiment analysis in IMDB and toxicity detection in the CivilComments Kill and CivilComments SS. 
%For CivilComments SS, we consider the 5\% overlap with 0.50 ATE dataset for evaluation.  

\noindent \textbf{Groups.} 
Classifiers that reduce spurious correlation are expected to decrease total accuracy but increase the accuracy of minority inputs that do not exhibit those correlations. To study such effects on accuracy, we divide our evaluation data into four groups: \texttt{Group1} $(Y\!=\!0,T\!=\!0)$, \texttt{Group2} $(Y\!=\!0,T\!=\!1)$, \texttt{Group3} $(Y\!=\!1,T\!=\!0)$, \texttt{Group4} 
$(Y\!=\!1,T\!=\!1)$. In addition, we report the average group accuracy across the four groups as a measure of debiasing/reduced spurious correlation. An ideal model should achieve both high overall accuracy and high average group accuracy, demonstrating its reduced reliance on spurious features. 
%Controlling the learnt feature effect helps us increase the accuracy on the minority groups while keeping the gains on the majority groups and overall accuracy. This also implies an increase in average of the four group accuracies. We hence highlight the gain in total accuracy while improving worst and average group accuracy

\noindent \textbf{Baselines.}
We consider popular baselines from prior work \cite{joshi2022all,he2022controlling,orgad2022debiasing}: 
weighting methods like DFL, DFL-nodemog, Product of Experts \cite{mahabadi2019end,orgad2022debiasing} and latent space removal methods like INLP \cite{ravfogel2020null}. We also include worst-group accuracy methods like GroupDRO, Subsampling \cite{sagawa2019distributionally,sagawa2020investigation} from the  machine learning literature, and a baseline \removetoken that removes the treatment feature from input (see Supp~\ref{sec:app_baselines}).

%  This is done by breaking the correlation between the treatment and the label in train distribution by Subsampling \cite{sagawa2020investigation} the data. GroupDRO \cite{sagawa2019distributionally} equally weighs both majority and minority groups. It additionally uses learnt groups wights and heavy regularisation to achieve best worst-group accuracy.

\noindent \textbf{Results.} %\todo{use Table. Tab seems too short.}
For the semi-synthetic dataset (CivilComments SS) in Table~\ref{tab:ceracc_ss_bert}, \cer{ate} increases the average group accuracy while retaining similar overall accuracy as \stdfine. \cer{ate} also has better minority group accuracy (i.e. \texttt{Group2},\texttt{Group3}) than \stdfine. In comparison, \cer{0} %regularises the feature effect to 0 (while the true effect is 0.50). This
leads to a decrease in overall accuracy and also average group accuracy compared to \cer{ate}.  Other baselines like \subsample, \groupdro or \dfl achieve a higher average group accuracy as they improve accuracy on the minority groups,  but they suffer a substantial reduction in overall accuracy, from 87 to 66-73, which hinders usability of the model. Methods like \dflsuccess or \poe have no impact  or obtain worse results compared to \stdfine.
%\removetoken presents an extreme version regularisation to 0 and hence shows major drop in overall accuracy. But decreasing the feature effect to 0 leads to better numbers on minority groups (at the expense of majority groups), and hence a better average accuracy.
These results show the fundamental tradeoff between total and average group accuracy and how \cer{ate} provides a good tradeoff between the two. 

%Other baselines like \subsample and \groupdro also lie on this tradeoff scale (reducing in-domain for better avg/worst group). Note how \removetoken is objectively better than both \subsample and \groupdro on average group and overall accuracy. \subsample,\groupdro might introduce new correlations into the data, which while helping worst/minority group accuracy leads to worse overall accuracy. \inlp along with removal of the biasing treatment token, also removes task critical information leading to worse total and avg accuracy. We can also see that \dfl is able to improve minority group accuracy but does so at cost of overall accuracy and is worse than \removetoken.
%\dflsuccess mimics the \stdfine in it's performance (as it is not able to learn main model's success probability). 

For the subsampled dataset (CivilComments Subsampled) in  Table~\ref{tab:ceracc_kill_bert}, we see a similar trend, where \cer{ate} gives the best tradeoff between overall and average accuracy. \cer{0} is substantially worse than \cer{ate}, showing the importance of not fully removing the effect of a spurious token. Except \poe, \subsample and \groupdro, all other methods obtain both lower total and average group accuracies compared to \cer{ate}. As before, \poe is near identical to \stdfine while the weighting methods \subsample and \groupdro  lead to significant decreases in total accuracy. %\removetoken has good average accuracy (with worse overall accuracy). \dflsuccess,\poe perform similar to the \stdfine method, while \dfl performs worse on all metrics. \subsample and \groupdro improve average numbers but decrease overall numbers.  

Finally, we show results for IMDB where the causal graph is unknown and our assumptions from Fig.~\ref{fig:cg2} may not be valid. Nonetheless  Table~\ref{tab:ceracc_imdb_bert} shows that both \cer{ate} and \cer{0} achieve better average group accuracy with slightly better total accuracy than the \stdfine model. Other baselines follow their usual trend: ML weighting baselines (\subsample, \groupdro) suffer reductions in total accuracy, \dfl and \poe methods are unable to improve average group accuracy substantially, and \inlp is worse for both total and average group accuracy. %Hence for data for which the causal graph is unknown, we suggest trying both \cer{0} and \cer{ate} and picking the best model accordingly.  
Besides BERT, results using DistilBERT as a base model show a similar trend (Supp~\ref{sec:app_distil_cer}). We also report \cer{propen} numbers in Supp~\ref{sec:app_distil_cer_propen}.

\subsection{Detecting Annotator bias }
\label{subsec:eval_annotator}
\begin{table}[h]
\centering
\resizebox{0.5\textwidth}{!}{%
\begin{tabular}{|l|Hll||l|Hll|}
\hline
Token & Direct  & Riesz DR & $P(Y|T)$ & 
Token & Direct  & Riesz DR & $P(Y|T)$ \\
\hline
gay & 21.02 $\pm$ 0.81 & 22.30 $\pm$ 1.03 & 0.66 & hate & 8.54 $\pm$ 2.06 & 5.81 $\pm$ 0.21 & 0.68 \\
racist & 10.71 $\pm$ 2.28 & 14.61 $\pm$ 0.97 & 0.75 & you're & 2.03 $\pm$ 0.30 & 1.99 $\pm$ 0.54 & 0.58\\
black & 14.66 $\pm$ 1.10 & 12.87 $\pm$ 0.36 & 0.69 & president & 1.01 $\pm$ 0.47 & 0.19 $\pm$ 0.21 & 0.55\\
white & 9.35 $\pm$ 1.12 & 9.91 $\pm$ 0.34 & 0.67 &guys & 0.87 $\pm$ 0.67 & 0.13 $\pm$ 1.24 & 0.58\\ 
\hline
\end{tabular}
}
\caption{\label{tab:annotator}
Tokens \texttt{racist} and \texttt{guys} show expected feature effect (1 and 0 resp.), but high feature effect for \texttt{black} and \texttt{gay} suggests annotator bias in dataset.}
% \vspace*{-15pt}
\end{table} 

% \begin{table}
% \centering
% \resizebox{0.5\textwidth}{!}{%
% \begin{tabular}{|l|Hll|}
% \hline
% Token & Direct  & Riesz DR & $P(Y|T)$  \\
% \hline
% % stupid & 39.81 $\pm$ 1.72 & 40.12 $\pm$ 0.78 & 0.94 \\
% gay & 21.02 $\pm$ 0.81 & 22.30 $\pm$ 1.03 & 0.66 \\
% racist & 10.71 $\pm$ 2.28 & 14.61 $\pm$ 0.97 & 0.75 \\
% % kill & 12.26 $\pm$ 0.98 & 13.31 $\pm$ 0.40 & 0.69 \\
% black & 14.66 $\pm$ 1.10 & 12.87 $\pm$ 0.36 & 0.69 \\
% % police & 9.09 $\pm$ 0.17 & 11.09 $\pm$ 0.31  & 0.55\\
% white & 9.35 $\pm$ 1.12 & 9.91 $\pm$ 0.34 & 0.67 \\ 
% % donald & 1.82 $\pm$ 1.43 & 6.33 $\pm$ 1.14 & 0.63\\
% hate & 8.54 $\pm$ 2.06 & 5.81 $\pm$ 0.21 & 0.68 \\
% % folks & 3.94 $\pm$ 0.62 & 5.11 $\pm$ 1.48 & 0.58\\
% trump & 6.33 $\pm$ 1.14 & 4.33 $\pm$ 0.33 & 0.62\\
% you're & 2.03 $\pm$ 0.30 & 1.99 $\pm$ 0.54 & 0.58\\
% % country & 0.35 $\pm$ 0.63 & 0.88 $\pm$ 0.34 & 0.52\\
% president & 1.01 $\pm$ 0.47 & 0.19 $\pm$ 0.21 & 0.55\\
% guys & 0.87 $\pm$ 0.67 & 0.13 $\pm$ 1.24 & 0.58\\
% \hline
% \end{tabular}
% }
% \caption{\label{tab:annotator}
% Annotator bias}
% \end{table} 

% \begin{table}
% \centering
% \resizebox{0.49\textwidth}{!}{%
% \begin{tabular}{|l|lllll|}
% \hline
% Method & white & black & president & you're & guys \\
% \hline
% % Riesz DR & 9.91 $\pm$ 0.34 & 12.87 $\pm$ 0.36 & 0.19 $\pm$ 0.21 & 1.99 $\pm$ 0.54 & 0.13 $\pm$ 1.24\\
% Riesz DR & 9.91 & 12.87  & 0.19  & 1.99  & 0.13 \\
% $\mathbb{E}[Y|T=1]$ & 0.67 & 0.69 & 0.55 & 0.58 & 0.58\\
% \hline
% \end{tabular}
% }
% \caption{\label{tab:annotator}
% Annotator bias}
% \end{table}

\vspace{-0.3cm}
% \label{subsec:app_annotator}
% While we focused on the debiasing task for classifiers, our feature effect estimator is general: we apply it to detect annotator bias in the  CivilComments dataset. If a token's true feature effect
% is known, we compare it to the estimated effect to detect annotator bias in the dataset.
While we focused on the debiasing task for classifiers, our feature effect estimator is general: we apply it to detect annotator bias in the  CivilComments dataset.
If the true feature effect of a token is known, we can compare it to the estimated effect to detect any annotator bias in the dataset. For tokens like ``racist'' and ``guys'' where the true effect is likely to be high and zero respectively, the estimated effect confirms the prior (see Table~\ref{tab:annotator}). But for tokens like ``gay'' or ``black'', our method shows a significant non-zero feature effect on the label which may indicate annotator bias, as it may be known that these tokens should have a zero effect on the toxicity label. 
Compared to the naive conditional probability ($Y|T$), our effect estimator can be used to  provide a better sense of how important certain keywords are for generating the output label.  (e.g., ``guys'' obtains a zero causal effect but $P(Y|T)$
shows a substantial deviation from 0.5). %Our feature effect estimator is a general method and can be used to detect annotator bias. 

\section{Conclusion}
Rather than fully removing a feature's effect on the classifier, we presented a method for fine-grained control of the feature's effect based on causal inference. We showed how our method allows a better tradeoff between overall accuracy and accuracy over subgroups in the data. Our preliminary study on annotator bias demonstrated that our method may be useful for detecting biases in the classification label too. As future work,  a natural direction is to combine these two threads and explore how we can develop methods to regularize features' effect  on the  debiased  label, rather than the (possibly confounded) labels provided in the dataset. %As future work, it will be useful to study how  features that cannot be   We presented a novel method for estimating effects of different features on the output label that works even at high levels of spurious correlations in data. We also presented an two-phase algorithm \cero which controls the learnt effect of features in the classifier to the estimated feature effect (or any other desired value) through automated augmentations. We showed how controlling classifier's learnt effect to the estimated feature effect leads to improvements on minority group accuracy and an increase in total accuracy compared to other strong baselines across tasks and datasets. We also include results for showing the utility of our method in detecting annotator bias, which also serves as an interesting future direction.

%\section{Conslusion}
%We presented a method for estimating feature effects and reducing learnt correlations in classifiers. 

% \input{files/cutout}

% \vspace{-0.2cm}
\paragraph{Limitations}
One major shortcoming of \cero method is the dependency on creation of counterfactual inputs. If there is an error in counterfactual generation, we might get a wrong feature effect estimate. Thus, for simplicity, our evaluation considered tokens as features. The parallel development of counterfactual input generation methods  \cite{wu2021polyjuice,howard2022neurocounterfactuals} would hopefully ease this issue and allow \cero to be used reliably for spurious correlations on more complex features too.

\vspace{-0.2cm}
\paragraph{Ethics Statement}
This project aims to check when methods are using spurious correlation. Identification of these spurious correlation is important for debiasing i.e. removal of dependence of the model on these correlations. Our work shows how instead of complete removal of these spurious features, regularising them might be better. At the same time, this is early research work and shouldn't be used in real-world systems without further evaluation. 

%\section*{Acknowledgements}

\bibliography{anthology,custom}
\bibliographystyle{acl_natbib}

\appendix

% \section{Appendix}
% \label{sec:appendix}

\section{Training Details}
\label{sec:app_training}

\paragraph{Architecture} All classification methods were trained using a single linear layer on top of BERT(/DistilBERT) [CLS] token. \riesz uses a common BERT model for sentence reprensentation and then uses 2 seperate linear layers for learning $\alpha_\rr$ and $g$ seperately. 

\paragraph{Seeds} We use three seeds for our experiments. 0,11,44. All numbers are reported with mean and std errors over these three seeds. 

\paragraph{Optimization} We use 1e-5 learning rate for BERT parameters and 1e-4 for the final linear layer parameters. We train with 32 batch size for all our experiments. The learning rate linearly decays over training iterations. We use Adam optimizer with 1e-2 weight decay for all methods.

\paragraph{Best Model Selection} All models are trained to completion (i.e. number of epochs specified for particular dataset). The evaluation is done after every epoch and the best model is chosen over all the epochs using the validation set. 

\paragraph{Loss} Binary cross entropy loss is used for all methods.

\paragraph{Tokenization} We use the standard uncased tokenizers with max length of 256 tokens.  

\section{Dataset Specific Details}

%%%%kill dataset

 %%%imdb

\label{sec:app_dataset}
For all datasets we set the number of epochs such that for all methods  the validation loss has bottomed and starts increasing. 

\paragraph{CivilComments Semi-Synthetic}  
Since CivilComments is heavily skewed towards the 0 label, we resample the dataset to create a balanced data which we use in all our experiments.
Since the  writer's intent (confounder) is unknown, we construct it as a property of the input text, $W=h(X) \in \{0,1\}$, leading to the modified causal graph in Fig.~\ref{fig:cg23}.
This property could be something simple like presence of a certain word like \texttt{police} in text or something more complex like inferred ethnicity of the writer.
Rather than choosing a property manually, we train distilbert for modeling $h(.)$ for a few hundred iterations. We hence use $W=h(X)$ as the property.  $h(.)$ achieves $\sim 78\%$ accuracy on the task. 
To ensure overlap, the treatment variable is sampled from $W$ such that $0<P(T|X)<1$ or equivalently $0<P(T|W)<1$. We do this by using $T$ equal to $W$ with $\epsilon>0$ fraction of samples flipped. 
Finally we sample a new label as $Y' \sim \textrm{Bernoulli}((1-\tau)Y + \tau T)$, giving the true feature effect as $\tau$. %The final graph can be seen in Fig~\ref{fig:cg3}. 
The complete text $Z=(X,T)$ is constructed by prepending each covariate sentence $X$ with the word \treated if $T=1$ and \untreated if $T=0$. 
This is true for all the experiments and datasets in our setup. This also eases counterfactual generation by just changing the prepended text from \treated to \untreated (and vice-versa). 
The dataset has 7K train samples and 2K test samples. We train the model for 10 epochs.
For controlling learnt effect, we use 0.50 ATE and 5\% overlap SS.

\paragraph{CivilComments Subsampled}  
Since \texttt{kill} doesn't occur often in dataset (3\%) we retain only 10\% of the untreated sentences. 

We subsample so as to retain only 5\% of the samples having $T=1 \& \ W=0$. Samples having $T=1,W=1$ are untouched. Samples having $T=0$ are subsampled by 10\% (as mentioned above). Our dataset has 5K train samples and 2K test samples. We train the model for 10 epochs.

\paragraph{IMDB} The dataset is subsampled to have equal number of positive and negative sentiment reviews. The \treated token is predictive of the sentiment with 90\% accuracy. The test set is constructed similarly. The dataset has 1354 train samples and 1328 test samples. We train the model for 30 epochs. 

\section{Method Specific Details}
\label{sec:app_baselines}

\paragraph{\cero} We use $\lambda = 0.1$ for our feature effect augmentation, i.e. loss on augmented samples is weighed 1e-1 times the loss on original samples.  

\paragraph{\subsample,\groupdro} These method considers an alternate objective of maximising worst group accuracy as a condition for learning models robust to spurious correlations. For Subsample we break the correlation between $T$ and $Y$ but maintain $P(T=1)$ and $P(Y=1)$ invariant (following \cite{joshi2022all}). i.e. for an input sample $P(T=1,Y=1) = P(T=1)P(Y=1)$. For GroupDRO we sample from all the four groups (as defined in Sec~\ref{subsec:cer_results}) equally, i.e. $P(T=1,Y=1) = 0.25$. Additionally we have corresponding groups weights (following the original paper) with step size of 0.01. We use heavy regularisation of 1e-2 with Adam optimizer (regularisation of 1e-1 led to degradation in numbers). 

\paragraph{\dfl,\poe,\dflsuccess} For training the biased/weak learner model we use TinyBERT model \footnote{\url{https://huggingface.co/prajjwal1/bert-tiny}}. The optimization parameters for TinyBERT model were same as that of the main model (described above). We observed that while \dfl and \poe's weak learner was able to capture the bias, \dflsuccess struggled to learn main model's success and collapsed to constant value. For \poe we use $\lambda = 1.0$, i.e. the loss minimised is $\textrm{CE}(f_m(X),Y) + \textrm{CE}(\textrm{Softmax}(\textrm{Log}(f_b(X))+\textrm{Log}(f_m(X))),Y)$

\paragraph{\inlp} We train \inlp in post-hoc fashion i.e we first train a \stdfine model, select the best model and then apply \inlp on its representation. We take the code from the official repository \footnote{\url{https://github.com/shauli-ravfogel/nullspace_projection}} and run it for 100 iterations with minimum accuracy stopping criterion of 0.50. We tried RLACE algorithm too, but it yeilded similar/worse results than \inlp

\section{Best Propensity and Riesz Eval}
\label{sec:app_bestpropensityeval}
\paragraph{Propensity Eval} We choose $\lambda=1.0$ as the best value from the table below. 

\begin{table}[h]
\centering
\resizebox{0.5\textwidth}{!}{%
\begin{tabular}{|l|lll|}
\hline
Dataset & $\lambda=0.1$ & $\lambda=1.0$ & $\lambda=10.0$ \\
\hline
1\% & 15.50 $\pm$ 0.32 & 13.62 $\pm$ 0.26 & 13.08 $\pm$ 0.31\\
5\% & 27.31 $\pm$ 0.02 & 25.29 $\pm$ 0.26 &  25.51 $\pm$ 0.39\\
10\% & 38.97 $\pm$ 0.19 & 36.20 $\pm$ 0.18 &  36.36 $\pm$ 0.14\\
\hline
\end{tabular}
}
\caption{\label{citation-guide}
Propensity validation loss for different hyperparameter $\lambda$. We choose $\lambda=1.0$ as the best value.}
\end{table}

% \section{Best Riesz Eval}
% \label{sec:app_bestrieszeval}
\paragraph{Riesz Eval} We choose $\lambda=0.01$ as the best value from the table below.

\begin{table}[h]
\centering
\resizebox{0.5\textwidth}{!}{%
\begin{tabular}{|l|lll|}
\hline
Dataset & $\lambda=0.01$ & $\lambda=0.1$ & $\lambda=1.0$ \\
\hline
1\% &  -9.71 $\pm$ 0.09 & -64.76 $\pm$ 3.72  & -68.74 $\pm$ 2.11\\
5\% &  -17.83 $\pm$ 0.20 & -17.87 $\pm$ 0.15  &  -17.28 $\pm$ 0.16\\
10\% & -61.42 $\pm$ 1.27 & -9.93 $\pm$ 0.11   & -9.38 $\pm$ 0.29\\
\hline
\end{tabular}
}
\caption{\label{citation-guide}
Riesz validation loss for different hyperparameter $\lambda$. We choose $\lambda=0.01$ as the best value.}
\end{table}

% \section{Riesz and Propensity Estimates for different hyperparameters}
% \label{sec:app_reiszpropenest}

% \begin{table*}[h]
% \centering
% \begin{tabular}{|l|lll|}
% \hline
% Method & 1\% Overlap & 5\% Overlap & 10\% Overlap  \\
% \hline
% Direct  & 11.30 $\pm$ 5.34 & 10.09 $\pm$ 7.44 & 2.48 $\pm$ 0.49  \\
% Propensity DR ($\lambda=0.1$) & 5.58 $\pm$ 3.44 & 9.80 $\pm$ 2.68 & 6.35 $\pm$ 0.55 \\
% Propensity DR ($\lambda=1.0$)& 37.42 $\pm$ 9.45 & 8.18 $\pm$ 4.78 & 1.26 $\pm$ 0.57 \\
% Propensity DR ($\lambda=10.0$) & 51.73 $\pm$ 20.65 & 5.27 $\pm$ 1.38 & 2.13 $\pm$ 1.23 \\
% Riesz DR ($\lambda=0.01$) & \textbf{4.50 $\pm$ 0.49} & \textbf{1.70 $\pm$ 1.12} & \textbf{ 1.17 $\pm$ 0.68} \\
% Riesz DR ($\lambda=0.1$) & 6.96 $\pm$ 1.46 & 1.87 $\pm$ 1.57 & 4.59 $\pm$ 0.40 \\
% Riesz DR ($\lambda=1.0$) & 19.33 $\pm$ 3.77 & 5.46 $\pm$ 2.21 & 6.06 $\pm$ 2.81 \\
 
% \hline
% \end{tabular}
% \caption{\label{citation-guide}
% Percentage Error in ATE Estimation with True ATE of 0.30 and DistilBERT Model}
% \end{table*}

\section{BERT Propensity-DR based FEAG numbers}
\label{sec:app_distil_cer_propen}

Propensity-DR based FEAG numbers on the three datasets are given in Table~\ref{tab:ceracc_ss_bert_feagpropen}, Table~\ref{tab:ceracc_kill_bert_feagpropen} and Table~\ref{tab:ceracc_imdb_bert_feagpropen}.

\begin{table*}[h]
\centering
\resizebox{\textwidth}{!}{%
\begin{tabular}{|l||llll|HHllH|}
\hline
Method & Group1 & Group2 & Group3 & Group4 & Treatment0 & Treatment1 & Total & Avg Group & Direct \\  
\hline
\cer{0} & 98.89 $\pm$ 0.48  & 7.48 $\pm$ 1.77  & 4.03 $\pm$ 1.53  & 97.40 $\pm$ 0.76  & 88.07 $\pm$ 0.29  & 85.99 $\pm$ 0.44  & 87.01 $\pm$ 0.34  & 51.95 $\pm$ 0.31  & 47.04 $\pm$ 3.88  \\
\cer{ate} & 98.30 $\pm$ 0.30  & 4.13 $\pm$ 0.94  & 7.75 $\pm$ 1.28  & 99.36 $\pm$ 0.18  & 87.97 $\pm$ 0.14  & 87.27 $\pm$ 0.06  & {87.62 $\pm$ 0.06}  & 52.39 $\pm$ 0.16  & 51.59 $\pm$ 0.18  \\
\cer{propen} & 100.00 $\pm$ 0.00 & 0.00 $\pm$ 0.00 & 0.00 $\pm$ 0.00 & 100.00 $\pm$ 0.00 & 88.59 $\pm$ 0.00 & 87.31 $\pm$ 0.00 & 87.94 $\pm$ 0.00 & 50.00 $\pm$ 0.00 & 83.33 $\pm$ 1.23 \\
\hline
\end{tabular}
}
\caption{\label{tab:ceracc_ss_bert_feagpropen}
Civil Comments Semi-Synthetic (0.50 ATE, 5\% overlap); models trained using BERT.}
\vspace{-1em}
\end{table*}

\begin{table*}[h]
\centering
\resizebox{\textwidth}{!}{%
\begin{tabular}{|l||llll|HHllH|}
\hline
Method & Group1 & Group2 & Group3 & Group4 & Treatment0 & Treatment1 & Total & Avg Group & Direct \\  
\hline
\cer{0} & 78.25 $\pm$ 0.45  & 11.59 $\pm$ 1.18  & 79.43 $\pm$ 0.25  & 94.25 $\pm$ 0.35  & 78.82 $\pm$ 0.16  & 78.98 $\pm$ 0.39  & 78.87 $\pm$ 0.14  & 65.88 $\pm$ 0.28  & 9.25 $\pm$ 0.42  \\
\cer{ate} & 78.80 $\pm$ 0.32  & 10.14 $\pm$ 0.59  & 80.34 $\pm$ 0.32  & 95.73 $\pm$ 0.35  & 79.55 $\pm$ 0.30  & 79.92 $\pm$ 0.19  & {79.66 $\pm$ 0.17}  & 66.25 $\pm$ 0.22  & 14.92 $\pm$ 0.19  \\
\cer{propen} & 77.60 $\pm$ 1.57 & 0.00 $\pm$ 0.00 & 77.93 $\pm$ 1.57 & 99.84 $\pm$ 0.23 & 77.76 $\pm$ 0.16 & 81.39 $\pm$ 0.19 & 78.83 $\pm$ 0.15 & 63.84 $\pm$ 0.12 & 29.04 $\pm$ 0.60 \\
\hline
\end{tabular}
}
\caption{\label{tab:ceracc_kill_bert_feagpropen}
CivilComments Subsampled dataset; models trained using BERT.}
\vspace{-1em}
\end{table*}

\begin{table*}[h]
\centering
\resizebox{\textwidth}{!}{%
\begin{tabular}{|l||llll|HHllH|}
\hline
Method & Group1 & Group2 & Group3 & Group4 & Treatment0 & Treatment1 & Total & Avg Group & Direct \\  
\hline
\cer{0} & 94.63 $\pm$ 0.72  & 33.33 $\pm$ 7.23  & 46.75 $\pm$ 1.84  & 97.30 $\pm$ 1.09  & 88.46 $\pm$ 0.47  & 90.23 $\pm$ 0.24  & {89.33 $\pm$ 0.15}  & 68.00 $\pm$ 1.65  & 44.84 $\pm$ 2.76  \\
\cer{ate} & 95.46 $\pm$ 1.27  & 15.34 $\pm$ 3.03  & 43.29 $\pm$ 5.49  & 99.34 $\pm$ 0.28  & 88.74 $\pm$ 0.43  & 90.06 $\pm$ 0.13  & {89.38 $\pm$ 0.16}  & 63.36 $\pm$ 1.75  & 57.26 $\pm$ 2.88  \\
\cer{propen} & 91.68 $\pm$ 2.20 & 39.15 $\pm$ 7.14 & 57.14 $\pm$ 2.81 & 96.84 $\pm$ 0.58 & 87.24 $\pm$ 1.58 & 90.47 $\pm$ 0.30 & 88.81 $\pm$ 0.68 & 71.21 $\pm$ 1.77 & 40.13 $\pm$ 0.49 \\
\hline
\end{tabular}
}
\caption{\label{tab:ceracc_imdb_bert_feagpropen}
IMDB dataset; models trained using BERT.}
\vspace{-1em}
\end{table*}

\section{DistilBERT FEAG numbers}
\label{sec:app_distil_cer}
We also show FEAG numbers on the three datasets using DistilBERT as the model in Table~\ref{tab:ceracc_ss_distil}, Table~\ref{tab:ceracc_kill_distil} and Table~\ref{tab:ceracc_imdb_distil}

\begin{table*}
\centering
\resizebox{\textwidth}{!}{%
\begin{tabular}{|l||llll|HHllH|}
\hline
Method & Group1 & Group2 & Group3 & Group4 & Treatment0 & Treatment1 & Total & Avg Group & Direct \\
\hline
\stdfine & 99.53 $\pm$ 0.20  & 3.96 $\pm$ 1.27  & 2.62 $\pm$ 1.37  & 99.50 $\pm$ 0.14  & 88.48 $\pm$ 0.06  & 87.37 $\pm$ 0.04  & 87.92 $\pm$ 0.03  & 51.40 $\pm$ 0.57  & 59.00 $\pm$ 3.42  \\
\removetoken & 91.53 $\pm$ 1.20  & 26.56 $\pm$ 3.00  & 26.28 $\pm$ 2.11  & 90.50 $\pm$ 1.14  & 84.09 $\pm$ 0.82  & 82.38 $\pm$ 0.62  & 83.23 $\pm$ 0.09  & 58.72 $\pm$ 0.24  & 0.26 $\pm$ 0.24  \\
\hline
\dfl & 83.86 $\pm$ 1.75  & 49.60 $\pm$ 4.03  & 35.05 $\pm$ 3.17  & 68.01 $\pm$ 3.35  & 78.29 $\pm$ 1.20  & 65.67 $\pm$ 2.41  & 71.89 $\pm$ 0.75  & 59.13 $\pm$ 0.20  & 2.32 $\pm$ 0.47  \\
\dflsuccess & 99.55 $\pm$ 0.17  & 2.99 $\pm$ 1.37  & 1.81 $\pm$ 0.62  & 99.58 $\pm$ 0.16  & 88.40 $\pm$ 0.08  & 87.32 $\pm$ 0.06  & 87.85 $\pm$ 0.02  & 50.98 $\pm$ 0.39  & 61.49 $\pm$ 3.03  \\
\poe & 99.99 $\pm$ 0.01  & 0.88 $\pm$ 0.72  & 0.00 $\pm$ 0.00  & 99.81 $\pm$ 0.16  & 88.58 $\pm$ 0.01  & 87.25 $\pm$ 0.05  & 87.91 $\pm$ 0.02  & 50.17 $\pm$ 0.14  & 70.53 $\pm$ 3.33  \\
\hline
\inlp & 99.78 $\pm$ 0.18  & 99.56 $\pm$ 0.36  & 0.60 $\pm$ 0.38  & 0.60 $\pm$ 0.47  & 88.47 $\pm$ 0.12  & 13.16 $\pm$ 0.37  & 50.28 $\pm$ 0.13  & 50.14 $\pm$ 0.08  & 51.97 $\pm$ 1.89  \\
\hline
\subsample & 74.50 $\pm$ 8.65  & 46.44 $\pm$ 12.78  & 45.52 $\pm$ 13.24  & 69.86 $\pm$ 12.15  & 71.19 $\pm$ 6.16  & 66.89 $\pm$ 8.99  & 69.01 $\pm$ 1.87  & 59.08 $\pm$ 1.05  & 5.38 $\pm$ 1.99  \\
\groupdro & 74.45 $\pm$ 2.92  & 65.35 $\pm$ 5.57  & 47.73 $\pm$ 5.79  & 57.52 $\pm$ 4.80  & 71.40 $\pm$ 1.94  & 58.52 $\pm$ 3.56  & 64.87 $\pm$ 1.20  & 61.26 $\pm$ 1.27  & 16.05 $\pm$ 4.54  \\
\hline
\cer{0} & 96.23 $\pm$ 0.13  & 13.54 $\pm$ 2.28  & 15.21 $\pm$ 0.43  & 97.11 $\pm$ 0.58  & 86.99 $\pm$ 0.08  & 86.50 $\pm$ 0.23  & 86.74 $\pm$ 0.08  & 55.52 $\pm$ 0.46  & 23.46 $\pm$ 1.51  \\
\cer{ate} & 99.00 $\pm$ 0.25  & 7.12 $\pm$ 0.21  & 4.93 $\pm$ 1.15  & 98.90 $\pm$ 0.05  & 88.27 $\pm$ 0.09  & 87.25 $\pm$ 0.02  & 87.75 $\pm$ 0.05  & 52.49 $\pm$ 0.25  & 50.21 $\pm$ 1.41  \\
\hline
\end{tabular}
}
\caption{\label{tab:ceracc_ss_distil}
Accuracy across groups for CivilComments Semi-Synthetic (0.50 ATE,5\% Overlap). All models are trained using DistilBERT model}

\end{table*}

\begin{table*}
\centering
\resizebox{\textwidth}{!}{%
\begin{tabular}{|l||llll|HHllH|}
\hline
Method & Group1 & Group2 & Group3 & Group4 & Treatment0 & Treatment1 & Total & Avg Group & Direct \\  
\hline
\stdfine & 96.23 $\pm$ 1.95  & 22.22 $\pm$ 7.14  & 32.03 $\pm$ 6.78  & 99.21 $\pm$ 0.34  & 87.96 $\pm$ 0.97  & 90.70 $\pm$ 0.51  & 89.30 $\pm$ 0.53  & 62.42 $\pm$ 2.81  & 66.42 $\pm$ 5.27  \\
\removetoken & 75.30 $\pm$ 4.08  & 69.31 $\pm$ 3.77  & 74.03 $\pm$ 1.62  & 76.59 $\pm$ 2.23  & 75.14 $\pm$ 3.35  & 75.79 $\pm$ 1.68  & 75.46 $\pm$ 1.21  & 73.81 $\pm$ 1.13  & 0.08 $\pm$ 0.03  \\
\hline
\dfl & 97.57 $\pm$ 1.23  & 8.99 $\pm$ 5.52  & 26.41 $\pm$ 10.90  & 99.54 $\pm$ 0.24  & 88.41 $\pm$ 0.48  & 89.53 $\pm$ 0.41  & 88.96 $\pm$ 0.33  & 58.13 $\pm$ 3.39  & 69.51 $\pm$ 6.57  \\
\dflsuccess & 94.31 $\pm$ 1.39  & 28.57 $\pm$ 2.70  & 41.99 $\pm$ 3.89  & 99.21 $\pm$ 0.25  & 87.57 $\pm$ 0.72  & 91.40 $\pm$ 0.14  & 89.44 $\pm$ 0.43  & 66.02 $\pm$ 0.41  & 55.77 $\pm$ 3.21  \\
\poe & 96.29 $\pm$ 1.00  & 19.05 $\pm$ 5.85  & 38.96 $\pm$ 5.85  & 99.67 $\pm$ 0.11  & 88.91 $\pm$ 0.32  & 90.76 $\pm$ 0.60  & 89.81 $\pm$ 0.43  & 63.49 $\pm$ 2.31  & 66.77 $\pm$ 4.52  \\
\hline
\inlp & 76.90 $\pm$ 14.35  & 71.96 $\pm$ 18.57  & 31.17 $\pm$ 18.42  & 25.12 $\pm$ 18.55  & 71.01 $\pm$ 10.13  & 30.29 $\pm$ 14.46  & 51.14 $\pm$ 2.03  & 51.29 $\pm$ 1.03  & 66.95 $\pm$ 3.82  \\
\hline
\subsample & 71.08 $\pm$ 1.47  & 68.78 $\pm$ 1.14  & 71.43 $\pm$ 1.23  & 77.65 $\pm$ 1.60  & 71.13 $\pm$ 1.44  & 76.67 $\pm$ 1.31  & 73.83 $\pm$ 1.34  & 72.23 $\pm$ 0.87  & 0.39 $\pm$ 0.10  \\
\groupdro & 74.98 $\pm$ 3.66  & 70.37 $\pm$ 3.12  & 73.16 $\pm$ 1.87  & 78.57 $\pm$ 2.53  & 74.75 $\pm$ 3.32  & 77.66 $\pm$ 1.91  & 76.17 $\pm$ 2.12  & 74.27 $\pm$ 1.00  & 2.57 $\pm$ 1.58  \\
\hline
\cer{0} & 91.94 $\pm$ 0.74  & 47.09 $\pm$ 1.14  & 55.84 $\pm$ 3.41  & 94.74 $\pm$ 0.57  & 87.29 $\pm$ 0.21  & 89.47 $\pm$ 0.38  & 88.36 $\pm$ 0.25  & 72.40 $\pm$ 0.76  & 34.40 $\pm$ 0.71  \\
\cer{ate} & 96.42 $\pm$ 0.42  & 30.69 $\pm$ 6.10  & 44.16 $\pm$ 2.81  & 98.09 $\pm$ 0.79  & 89.69 $\pm$ 0.12  & 90.64 $\pm$ 0.25  & 90.15 $\pm$ 0.07  & 67.34 $\pm$ 0.84  & 52.65 $\pm$ 1.71  \\
\hline
\end{tabular}
}
\caption{\label{tab:ceracc_imdb_distil}
IMDB dataset; models trained using DistilBERT
}
\end{table*}

\begin{table*}
\centering
\resizebox{\textwidth}{!}{%
\begin{tabular}{|l||llll|HHllH|}
\hline
Method & Group1 & Group2 & Group3 & Group4 & Treatment0 & Treatment1 & Total & Avg Group & Direct \\  
\hline
\stdfine & 80.22 $\pm$ 0.58  & 5.80 $\pm$ 0.59  & 76.32 $\pm$ 0.47  & 97.70 $\pm$ 0.35  & 78.32 $\pm$ 0.08  & 80.72 $\pm$ 0.38  & 79.03 $\pm$ 0.06  & 65.01 $\pm$ 0.19  & 19.07 $\pm$ 0.39  \\
\removetoken & 76.72 $\pm$ 0.68  & 12.32 $\pm$ 0.59  & 84.02 $\pm$ 0.25  & 90.31 $\pm$ 0.97  & 80.28 $\pm$ 0.28  & 75.90 $\pm$ 0.68  & 78.99 $\pm$ 0.36  & 65.84 $\pm$ 0.20  & 10.03 $\pm$ 0.49  \\
\hline
\dfl & 85.57 $\pm$ 1.63  & 8.70 $\pm$ 2.72  & 67.01 $\pm$ 1.94  & 93.60 $\pm$ 0.70  & 76.53 $\pm$ 0.51  & 77.91 $\pm$ 0.68  & 76.94 $\pm$ 0.56  & 63.72 $\pm$ 0.86  & 17.87 $\pm$ 0.42  \\
\dflsuccess & 77.27 $\pm$ 3.18  & 0.00 $\pm$ 0.00  & 77.59 $\pm$ 2.54  & 98.69 $\pm$ 0.49  & 77.42 $\pm$ 0.40  & 80.46 $\pm$ 0.39  & 78.32 $\pm$ 0.20  & 63.39 $\pm$ 0.08  & 19.15 $\pm$ 0.61  \\
\poe & 81.53 $\pm$ 0.91  & 16.67 $\pm$ 2.37  & 78.74 $\pm$ 0.09  & 93.60 $\pm$ 1.53  & 80.17 $\pm$ 0.42  & 79.38 $\pm$ 0.86  & 79.94 $\pm$ 0.12  & 67.63 $\pm$ 0.45  & 15.70 $\pm$ 0.80  \\
\hline
\inlp & 72.90 $\pm$ 1.55  & 10.87 $\pm$ 2.72  & 81.84 $\pm$ 1.08  & 91.46 $\pm$ 1.10  & 77.25 $\pm$ 0.28  & 76.57 $\pm$ 0.39  & 77.05 $\pm$ 0.13  & 64.27 $\pm$ 0.51  & 18.73 $\pm$ 0.75  \\
\hline
\subsample & 76.61 $\pm$ 1.29  & 39.13 $\pm$ 2.05  & 81.61 $\pm$ 0.82  & 81.28 $\pm$ 1.42  & 79.05 $\pm$ 0.33  & 73.49 $\pm$ 0.83  & 77.41 $\pm$ 0.31  & 69.66 $\pm$ 0.40  & 1.56 $\pm$ 0.36  \\
\groupdro & 78.14 $\pm$ 0.18  & 48.55 $\pm$ 3.88  & 77.47 $\pm$ 0.77  & 74.06 $\pm$ 1.19  & 77.82 $\pm$ 0.44  & 69.34 $\pm$ 0.29  & 75.32 $\pm$ 0.39  & 69.55 $\pm$ 0.47  & 0.52 $\pm$ 0.75  \\
\hline
\cer{0} & 77.70 $\pm$ 1.49  & 10.14 $\pm$ 1.57  & 78.62 $\pm$ 1.17  & 94.91 $\pm$ 0.94  & 78.15 $\pm$ 0.35  & 79.25 $\pm$ 0.58  & 78.48 $\pm$ 0.09  & 65.35 $\pm$ 0.25  & 10.19 $\pm$ 0.10  \\
\cer{ate} & 79.13 $\pm$ 0.85  & 9.52 $\pm$ 1.77  & 79.08 $\pm$ 1.32  & 96.72 $\pm$ 0.35  & 79.10 $\pm$ 0.25  & 80.05 $\pm$ 0.11  & 79.38 $\pm$ 0.15  & 66.36 $\pm$ 0.28  & 14.64 $\pm$ 0.68  \\
\hline
\end{tabular}
}
\caption{\label{tab:ceracc_kill_distil}
Accuracy across groups for CivilComments Subsampled trained using DistilBERT model.}
\end{table*}

\section{Alternative Causal Graphs}
\label{sec:alternate_causal_graphs}
\begin{figure}[h]
\begin{subfigure}{0.5\textwidth}
\includegraphics[width=\textwidth]{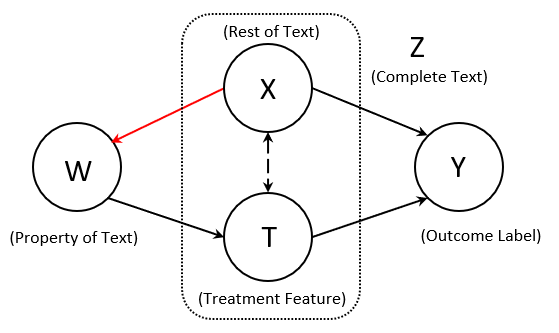}
\caption{Alternate Causal Graph. The Red arrow is different from Fig.~\ref{fig:cg}. The confounding now instead of being the intent of writer, is a property of text $P$}
\label{fig:cg2}
\end{subfigure}

\begin{subfigure}{0.5\textwidth}
\includegraphics[width=\textwidth]{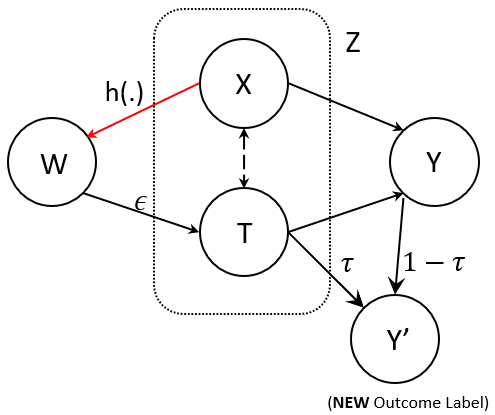}
\caption{Causal graph from Semi-Synthetic setting. The graph is derived from Fig.~\ref{fig:cg2}, with addition of new node $Y'$. The function $h(.)$ is used to get property $W$ from $X$. Noise $\epsilon$ (in form of label flipping) is added to $W$ to ensure non-zero $P(T=t|X)$, i.e. every co-variate $X$ has non-zero probability of being treated and being untreated. $\tau$ is $T$'s effect on the new outcome label $Y'$, while $1-\tau$ is $Y$'s affect on $Y'$}
\label{fig:cg3}
\end{subfigure}
\caption{Alternate Causal Graphs}
\label{fig:cg23}
\end{figure}

We present alternate version of the primary causal graph (Fig~\ref{fig:cg}) in Fig~\ref{fig:cg23}

\section{Label Flipping Algorithm}
\label{sec:app_label_flip}
Consider treatment $T$, label $Y$. The desired effect as $\tau$. WLOG we can assume $\tau>0$ (if $\tau<0$, then make $T' = 1-T$ and proceed with $T'$). The new counterfacutal labels are $Y^C$ and new treatment is $T^C=1-T$ (we will only use $T$ and $T^C$ will implicitly be $1-T$)

Consider probabilities as : 
\begin{equation}
    \begin{split}
        P(Y=1|T=1) & = p_1 \\
        P(Y=0|T=1) & = 1-p_1 \\
        P(Y=0|T=0) & = p_2 \\
        P(Y=1|T=0) & = 1-p_2 \\
    \end{split}
\end{equation}

\paragraph{Going from untreated to treated} 
Since $\tau>0$, changing treatment from 0 to 1, should increase the probability of outcome label being 1 (and decrease probability of it being 0) i.e. $P(Y^C=1|T=0)>(Y=1|T=0) \& P(Y^C=0|T=0)<(Y=0|T=0)$. 
This can be achieved by keeping $Y^C = Y$ whenever $Y=1$ and randomly flipping certain fraction (say $\eta$) of samples having $Y=0$ to $Y^C=1$ ( the other $1-\eta$ would have $Y^C=Y=0$)
With the goal of $P(Y^C=1|T=0)-P(Y=1|T=0) = \tau$, $\eta$ can be easily computed as $\frac{\tau}{p_2}$. To verify we can compute 
\begin{equation}
    \begin{split}
        P(Y^C=1|T=0) &= P(Y=1|T=0) + \\
                     & \eta P(Y=0|T=0) \\
        P(Y^C=1|T=0) &= P(Y=1|T=0) + (\frac{\tau}{p_2})p_2 \\
        P(Y^C=1|T=0) &-P(Y=1|T=0) =  \tau \\
    \end{split}
\end{equation}

\paragraph{Going from treated to untreated}
Similarly we can argue that $Y^C = Y$ whenever $Y=0$ and randomly flipping $\frac{\tau}{p_2}$ fraction of samples having $Y=1$ to $Y^C=0$.

\section{Computational Budget}

\paragraph{GPUs used} We run our experiments on NVIDIA RTX A6000 gpus. On an average each experiment takes 1 hour to complete. 

We use the BERT-base (110 Million parameters) and DistilBERT model (55 Million parameters) for computation.

\section{Two-Head Riesz Model}
\label{subsec:app_twohead}

\begin{figure}[h]
\includegraphics[width=0.45\textwidth]{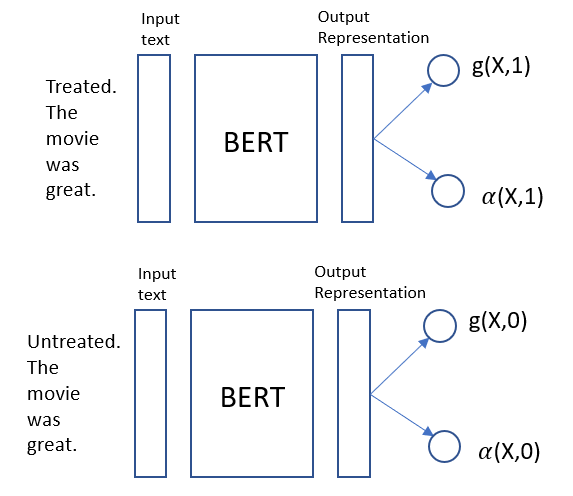}
\caption{Two-head model for jointly training $\alpha_\rr$ and $g$ for Riesz estimator. The top figure has a treated sentence as input while the bottom figure has an untreated sentence as input.}
\label{fig:riesz}
\end{figure}

Sharing parameters between classifier and Riesz estimator using a two-headed model forces the shared model (e.g. BERT) to learn representations which are important for both classifier and Riesz model. While this may cause a decrease in either model's performance, this leads to a better estimate due to reduced noise in estimation \cite{shi2019adapting}. We present our architecture in Fig~\ref{fig:riesz}

\end{document}